\documentclass[10pt,twocolumn,letterpaper]{article}

\usepackage[final]{cvpr}              

\usepackage[accsupp]{axessibility}
\usepackage{graphicx}
\usepackage{amssymb}
\usepackage{booktabs}
\usepackage{multicol}
\usepackage{multirow}
\usepackage{amsfonts}       
\usepackage{nicefrac}       
\usepackage{microtype}      
\usepackage{xspace}
\usepackage{graphicx}
\usepackage{epsfig}
\usepackage{cases}
\usepackage{color}
\usepackage{algorithm}
\usepackage{algorithmic}
\usepackage{tablefootnote,footnotehyper}
\usepackage{array}
\usepackage[dvipsnames]{xcolor}
\usepackage{amsmath}
\newcolumntype{L}[1]{>{\raggedright\arraybackslash}p{#1}}
\newcolumntype{C}[1]{>{\centering\arraybackslash}p{#1}}
\newcolumntype{R}[1]{>{\raggedleft\arraybackslash}p{#1}}

\usepackage[pagebackref,breaklinks,colorlinks]{hyperref}

\usepackage[capitalize]{cleveref}
\crefname{section}{Sec.}{Secs.}
\Crefname{section}{Section}{Sections}
\Crefname{table}{Table}{Tables}
\crefname{table}{Tab.}{Tabs.}
\newcommand{\red}[0]{\textcolor{red}}

\def\cvprPaperID{6414} 
\def\confName{CVPR}
\def\confYear{2023}

\begin{document}

\title{Improved Test-Time Adaptation for Domain Generalization}

\author{Liang Chen$^1$\quad 
Yong Zhang$^{2}$\thanks{Corresponding  authors. This work is done when L. Chen is an intern in Tencent AI Lab.}\quad 
Yibing Song$^3$\quad 
Ying Shan$^{2}$\quad 
Lingqiao Liu$^{1\ast}$\\
 {$^1$~The University of Adelaide}\quad  {$^2$~Tencent AI Lab}\quad
 {$^3$~AI$^3$ Institute, Fudan University}\\
 {\tt\small \{liangchen527, zhangyong201303, yibingsong.cv\}@gmail.com} \\ {\tt\small yingsshan@tencent.com}~~~~~{\tt\small lingqiao.liu@adelaide.edu.au}
}

\maketitle
\thispagestyle{empty}

\begin{abstract}
The main challenge in domain generalization (DG) is to handle the distribution shift problem that lies between the training and test data. Recent studies suggest that test-time training (TTT), which adapts the learned model with test data, might be a promising solution to the problem. Generally, a TTT strategy hinges its performance on two main factors: selecting an appropriate auxiliary TTT task for updating and identifying reliable parameters to update during the test phase. Both previous arts and our experiments indicate that TTT may not improve but be detrimental to the learned model if those two factors are not properly considered. This work addresses those two factors by proposing an \textbf{I}mproved \textbf{T}est-\textbf{T}ime \textbf{A}daptation~(ITTA) method. First, instead of heuristically defining an auxiliary objective, we propose a learnable consistency loss for the TTT task, which contains learnable parameters that can be adjusted toward better alignment between our TTT task and the main prediction task. Second, we introduce additional adaptive parameters for the trained model, and we suggest only updating the adaptive parameters during the test phase. Through extensive experiments, we show that the proposed two strategies are beneficial for the learned model (see Figure~\ref{fig tease}), and ITTA could achieve superior performance to the current state-of-the-art methods on several DG benchmarks. Code is available at \url{https://github.com/liangchen527/ITTA}.
\end{abstract}

\section{Introduction}
Recent years have witnessed the rapid development of deep learning models, which often assume the training and test data are from the same domain and follow the same distribution. However, this assumption does not always hold in real-world scenarios. Distribution shift among the source and target domains is ubiquitous in related areas \cite{koh2021wilds}, such as autonomous driving or object recognition tasks, resulting in poor performances for delicately designed models and hindering the further application of deep learning techniques.

Domain generalization (DG) \cite{muandet2013domain,ghifary2016scatter, li2018domain,hu2020domain,ganin2016domain,li2018domain,yang2021adversarial,li2018deep,li2018learning,balaji2018metareg,dou2019domain,li2019episodic,rame2021ishr,pezeshki2021gradient,chen2022comen}, designed to generalize a learned model to unseen target domains, has attracted a great deal of attention in the research community. The problem can be traced back to a decade ago \cite{blanchard2011generalizing}, and various approaches have been proposed to push the DG boundary ever since. Those efforts include invariant representation learning \cite{muandet2013domain,shi2021gradient,pandey2021generalization,harary2022unsupervised}, adversarial learning \cite{ganin2016domain,li2018domain,yang2021adversarial,li2018deep}, augmentation \cite{zhou2021domain,li2022uncertainty,xu2021fourier,li2021simple,chen2022mix}, or meta-learning \cite{li2018learning,balaji2018metareg,dou2019domain,li2019episodic}. Despite successes on certain occasions, a recent study \cite{gulrajani2020search} shows that, under a rigorous evaluation protocol, most of these arts are inferior to the baseline empirical risk minimization (ERM) method \cite{vapnik1999nature}. This finding is not surprising, as most current arts strive to decrease the distribution shift only through the training data while overlooking the contributions from test samples. 

\def\swone{0.95\linewidth}
\begin{figure}
    \centering
    \begin{tabular}{c}
    \centering
    \includegraphics[width=\swone]{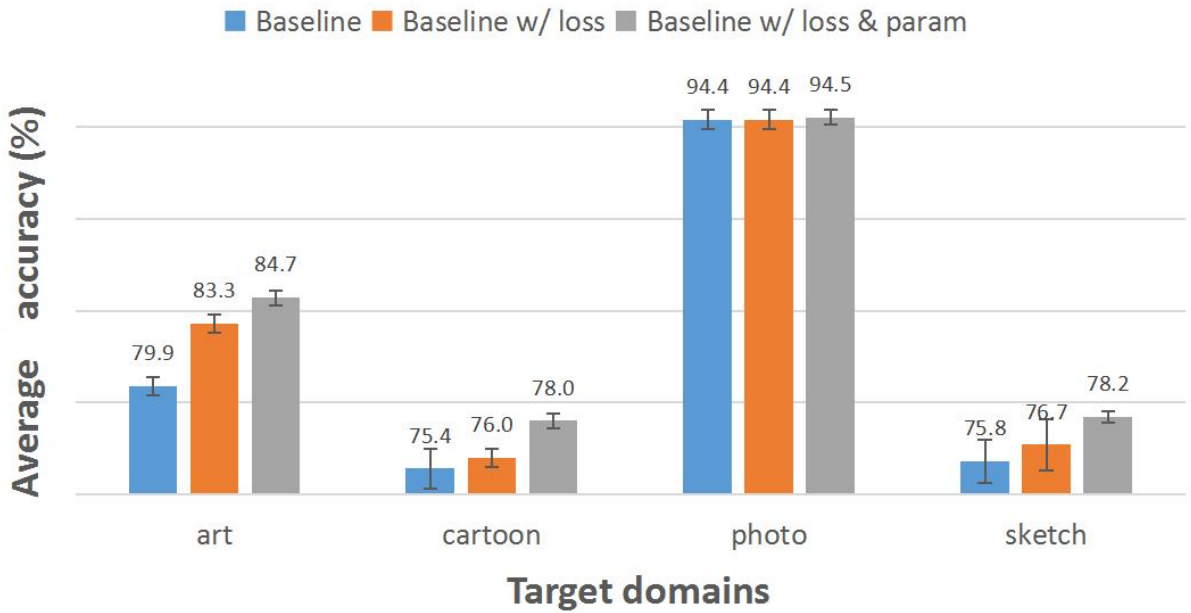}
    \end{tabular}
    \vspace{-0.3 cm}
    \caption{Performance improvements from the proposed two strategies (\ie introducing a learnable consistency \textbf{loss} and including additional adaptive \textbf{param}eters to improve TTT) for the baseline model (\ie ResNet18 \cite{he2016deep} with existing augmentation strategy \cite{zhou2021domain}). Experiments are conducted on the PACS dataset \cite{li2017deeper} with the leave-one-out setting. Following \cite{gulrajani2020search}, we use 60 sets of random seeds and hyper-parameters for each target domain. The reported average accuracy and error bars verify the effectiveness of our method.}
    \label{fig tease}
    \vspace{-0.5 cm}
\end{figure}

Recently, the test-time training (TTT) technique \cite{sun2020test} has been gaining momentum for easing the distribution shift problem. TTT lies its success in enabling dynamic tuning of the pretrained model with the test samples via an auxiliary TTT task, which seems to be a promising effort when confronting data from different domains. 
However, TTT is not guaranteed to improve the performance. Previous arts \cite{liu2021ttt++,wang2020tent} indicate that selecting an appropriate auxiliary TTT task is crucial, and an inappropriate one that does not align with the main loss may deteriorate instead of improving the performance. Meanwhile, it is pointed out in \cite{wang2020tent} that identifying reliable parameters to update is also essential for generalization, which is in line with our experimental findings in Sec.~\ref{sec param}. Both of these two tasks are non-trivial, and there are limited efforts made to address them.

This paper aims to improve the TTT strategy for better DG. First, different from previous works that empirically define auxiliary objectives and assume they are aligned with the main task, our work does not make such assumptions. Instead, we suggest learning an appropriate auxiliary loss for test-time updating. Specifically, encouraged by recent successes in multi-view consistency learning \cite{he2020momentum, chen2020simple, grill2020bootstrap}, we propose to augment the consistency loss by adding learnable parameters based on the original implementation, where the parameters can be adjusted to assure our TTT task can be more aligned with the main task and are updated by enforcing the two tasks share the same optimization direction.
Second, considering that identifying reliable parameters to update is an everlasting job given the growing size of current deep models, we suggest introducing new adaptive parameters after each block during the test phase, and we only tune the new parameters by the learned consistency loss while leaving the original parameters unchanged. 
Through extensive evaluations on the current benchmark \cite{gulrajani2020search}, we illustrate that the learnable consistency loss performs more effectively than the self-supervised TTT tasks adopted in previous arts \cite{sun2020test,wang2020tent}, and by tuning only the new adaptive parameters, our method is superior to existing strategies that update all the parameters or part of them.

This work aims to ease the distribution shift problem by improving TTT, and the main contributions are three-fold:
\begin{itemize}
    \item We introduce a learnable consistency loss for test-time adaptation, which can be enforced to be more aligned with the main loss by tuning its learnable parameters.
    \item We introduce new adaptive parameters for the trained model and only update them during the test phase.
    \item We conduct experiments on various DG benchmarks and illustrate that our ITTA performs competitively against current arts under the rigorous setting \cite{gulrajani2020search} for both the multi-source and single-source DG tasks. 
\end{itemize}

\section{Related Works}

\subsection{Domain Generalization.}
Being able to generalize to new environments while deploying is a challenging and practical requirement for current deep models. Existing DG approaches can be roughly categorized into three types. \textbf{(1) Invariant representation learning:} The pioneering work \cite{ben2006analysis} theoretically proves that if the features remain invariant across different domains, then they are general and transferable to different domains. Guided by this finding, \cite{muandet2013domain} uses maximum mean discrepancy (MMD) to align the learned features, and \cite{ghifary2015domain} proposes to use a multi-domain reconstruction auto-encoder to obtain invariant features. More recently, \cite{shi2021gradient} suggests maximizing the inner product of gradients from different domains to enforce invariance, and a similar idea is proposed in \cite{rame2021ishr} where these gradients are expected to be similar to their mean values. 
\textbf{(2) Optimization algorithms:} Among the different optimization techniques adopted in DG, prevailing approaches resort to adversarial learning \cite{ganin2016domain,li2018domain,yang2021adversarial,li2018deep} and meta-learning \cite{li2018learning,balaji2018metareg,dou2019domain,li2019episodic}. Adversarial training is often used to enforce the learned features to be agnostic about the domain information. In \cite{ganin2016domain}, a domain-adversarial neural network (DANN) is implemented by asking the mainstream feature to maximize the domain classification loss. This idea is also adopted in \cite{li2018deep}, where adversarial training and an MMD constraint are employed to update an auto-encoder. Meanwhile, the meta-learning technique is used to simulate the distribution shifts between seen and unseen environments \cite{li2018learning,balaji2018metareg,dou2019domain,li2019episodic}, and most of these works are developed based on the MAML framework \cite{finn2017model}.
\textbf{(3) Augmentation:} Most augmentation skills applied in the generalization tasks are operated in the feature level \cite{li2021simple,zhou2021domain,nam2021reducing,kim2021selfreg} except for \cite{yan2020improve,xu2021fourier, chen2022self} which mix images \cite{yan2020improve} or its phase \cite{xu2021fourier} to synthesize new data. To enable contrastive learning, we incorporate an existing augmentation strategy \cite{zhou2021domain} in our framework. This method originated from AdaIN \cite{huang2017arbitrary}, which synthesizes new domain information by mixing the statistics of the features. Similar ideas can be found in \cite{nam2021reducing,li2022uncertainty}.


\subsection{Test-Time Training and Adaptation}
Test-Time Training (TTT) is first introduced in \cite{sun2020test}. The basic paradigm is to employ a test-time task besides the main task during the training phase and update the pretrained model using the test data with only the test-time objective before the final prediction step. The idea is empirically proved effective \cite{sun2020test} and further developed in other related areas~\cite{wang2020tent,schneider2020improving,li2021test,bartler2022mt3,fleuret2021uncertainty,zhang2021adaptive,chen2022ost,choi2022improving,chen2022contrastive,gandelsman2022test,zhong2022meta,xiao2022learning}. Most current works focus on finding auxiliary tasks for updating during the test phase, and the efforts derive from self-supervion~\cite{sun2020test,bartler2022mt3,li2021test,chen2022contrastive,fleuret2021uncertainty,gandelsman2022test}, meta-learning~\cite{zhong2022meta,zhang2021adaptive,xiao2022learning}, information entropy~\cite{wang2020tent}, pseudo-labeling~\cite{chen2022ost,choi2022improving}, to name a few.
%
However, not all empirically selected test-time tasks are effective. A recent study~\cite{liu2021ttt++} indicates that only when the auxiliary loss aligns with the main loss can TTT improve the trained model. Inspired by that, we propose a learnable consistency loss and enforce alignment between the two objectives. Results show that our strategy can be beneficial for the trained model (see Figure~\ref{fig tease}).    

Meanwhile, \cite{wang2020tent} suggests that auxiliary loss is not the only factor that affects the performance. Selecting reliable parameters to update is also crucial within the TTT framework. Given the large size of current models, correctly identifying these parameters may require tremendous amounts of effort. To this end, instead of heuristically selecting candidates, we propose to include new adaptive parameters for updating during the test phase. Experimental results show that the proposed method can obtain comparable performances against existing skills.

\def\swone{0.95\linewidth}
\begin{figure}
    \centering
    \begin{tabular}{c}
    \centering
    \includegraphics[width=\swone]{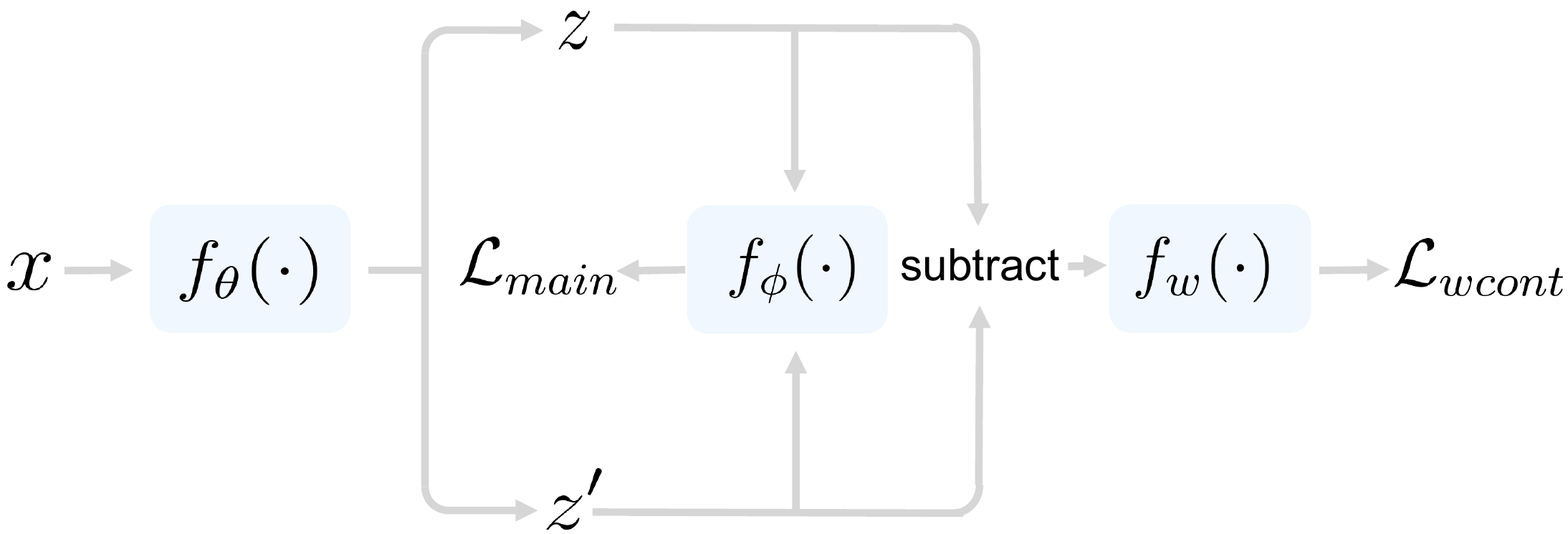}
    \end{tabular}
    \vspace{-0.3cm}
    \caption{Training process of ITTA. We use $x$ from the source domain as input for the feature extractor $f_{\theta}(\cdot)$ to obtain the representation $z$ and its augmented version $z'$, where the augmentation skill from \cite{zhou2021domain} is applied. The classifier $f_{\phi}(\cdot)$ and weight subnetwork $f_w(\cdot)$ are used to compute the main loss $\mathcal{L}_{main}$ and learnable consistency loss $\mathcal{L}_{wcont}$. Please refer to our text for details.}
    \label{fig pipeline}
    \vspace{-0.5 cm}
\end{figure}

\section{Methodology}
In the task of DG, we are often given access to data from $S$ ($S \geq 1$) source domains $\mathcal{D}_s = \{D_1, D_2,..., D_S\}$ and expect a model to make good prediction on unseen target domains $\mathcal{D}_t = \{D_1, D_2,...,D_T\}$ ($T \geq 1$).
Our method aims to improve the test-time training (TTT) strategy for better DG. The improvements are two-fold. First, we propose a learnable consistency loss for the TTT task, which could be enforced to align with the main objective by tuning its learnable weights. Second, we suggest including additional adaptive parameters and only updating these adaptive parameters during the test phase.

\newcommand{\D}{\hspace{0.4 cm}}
\newcommand{\comment}{\textcolor{SkyBlue}}
\begin{algorithm}[t]
\caption{Pseudo code of the training phase of ITTA in a PyTorch-like style.}
\footnotesize
\comment{\# ~~ $f_{\theta}, f_{\phi}, f_{w}$: ~feature extractor, classifier, weight subnetwork}\\
\comment{\# ~~ $\alpha$, $\textbf{0}$: ~weight paramter, all zero tensor} \vspace{0.3 cm}\\
\comment{\# ~~training process}\\
for $x, y$ in training\_loader: ~~\comment{\# ~~load a minibatch with N samples}\\
\begin{algorithmic}[]
\vspace{-0.4 cm}
\STATE def forward\_process($x,y$):\\
\STATE \D $z,z' = f_{\theta}.$forward($x$)\\
\STATE \D \comment{\# ~~computing losses}\\
\STATE \D $\mathcal{L}_{main}=$ CrossEntropyLoss($f_{\phi}.\text{forward}(z),~~y$) \\ 
\STATE \D $\mathcal{L}_{main}+=$ CrossEntropyLoss($f_{\phi}.\text{forward}(z'),~~y$)\\
\STATE \D $\mathcal{L}_{wcont}=$ MSELoss($f_w.\text{forward}(z-z')$,~~\textbf{0}) \vspace{0.3CM}\\
\STATE \D return $\mathcal{L}_{main},~~\mathcal{L}_{wcont}$ \vspace{0.3cm}\\
\STATE \comment{\# ~~SGD update: feature extractor and classifier}\\
\STATE $\mathcal{L}_{main},~~ \mathcal{L}_{wcont}$ = forward\_process($x,y$) \\
\STATE $([f_{\theta}$.params, $f_{\phi}$.params]).zero\_grad()
\STATE $(\mathcal{L}_{main}+\alpha \mathcal{L}_{wcont}).$backward() \\
\STATE update($\left[f_{\theta}.\text{params},~~ f_{\phi}.\text{params}\right]$) \vspace{0.3 cm}\\
\STATE \comment{\# ~~compute objectives for updating weight subnetwork}\\
\STATE $\mathcal{L}_{main},~~ \mathcal{L}_{wcont}$ = forward\_process($x,y$) \\
\STATE $\mathcal{L}_{main}$.backward()
\STATE $\hat{\textbf{g}}_{main}$ = $f_{\theta}$.params.grad.clone().normalize()
\STATE $f_{\theta}$.params.zero\_grad()
\STATE $\mathcal{L}_{wcont}$.backward()
\STATE $\hat{\textbf{g}}_{wcont}$ = $f_{\theta}$.params.grad.clone().normalize() \vspace{0.3cm}\\
\STATE \comment{\# ~~SGD update: weight subnetwork}\\
\STATE MSELoss($\hat{\textbf{g}}_{main}$,~~$\hat{\textbf{g}}_{wcont}$).backward()
\STATE $f_w$.params.zero\_grad()
\STATE update($f_w$.params)
\end{algorithmic}
\label{alg 1}
\end{algorithm}

\subsection{A Learnable Consistency Loss for TTT}
The TTT strategies have shown promising performances when dealing with distribution shift problems \cite{wang2020tent,li2021test}. However, their successes are depended on the empirically selected auxiliary TTT tasks, which may deteriorate the performances if chosen improperly. Motivated by the recent successes in multi-view consistency learning \cite{he2020momentum, chen2020simple, grill2020bootstrap}, we suggest adopting a consistency loss in our TTT task. Note that the naive consistency loss is still not guaranteed to be effective as prior art \cite{liu2021ttt++} indicates that only when the auxiliary loss aligns with the main loss, can TTT improves the performance. To this end, we propose to augment the auxiliary loss with learnable parameters that could be adjusted toward a better alignment between the TTT and main tasks. In our case, we make the adopted consistency loss learnable by introducing a weight subnetwork that allows flexible ways to measure the consistency between two views of the same instance.

We first introduce the pipeline of our training framework. Given the $D$ dimensional representation $z\in\mathbb{R}^{D}$\footnote{We omit the batch dimensions of the variables for simplicity.} and its corresponding augmented version $z'$ that are obtained from a feature extractor (\ie $\{z, z'\} = f_{\theta}(x)$, where $x$ is an input image from $\mathcal{D}_s$, and $f_{\theta}(\cdot)$ is the feature extractor parameterized by $\theta$. In our implementation, we use the existing augmentation method \cite{zhou2021domain} to obtain $z'$ by modifying the intermediate activation in $f_{\theta}(x)$. We show in our supplementary material that our framework can also thrive with other augmentation strategies), our learnable consistency loss is given by,
\begin{equation}
\label{eq wcloss}
\mathcal{L}_{wcont} = \Vert f_w(z - z')\Vert,
\end{equation}
where $\Vert\cdot\Vert$ denotes the $L2$ norm; $f_w(\cdot)$ is the weight subnetwork parameterized by $w$.
To make the training process more stable and potentially achieve better performance, we apply a dimension-wise nonlinear function to map each dimension of $z-z'$ before calculating the $L2$ norm. That is, $\forall h \in \mathbb{R}^D$, $f_w(h)$ is implemented by stacking layers of a nonlinear function: $\text{ReLU}(a*h + b)$, where $a \in \mathbb{R}^D$ and $b \in \mathbb{R}^D$ are the weight and bias from the nonlinear function, and different layers of $a, b$ form the parameter $w$ in $f_w$. In effect, this creates a piecewise-linear mapping function for $h$: depending on the value of $h$, the output could be 0, a constant, or a scaling-and-shifted version of $h$. More studies about the design of $f_w$ are provided in our supplementary material. 
Compared to the naive consistency learning without $f_w$, our $\mathcal{L}_{wcont}$ can be more flexible with an adjustable $f_w$, which we show in the following is the key for learning an appropriate loss in the improved TTT framework.

\def\swone{0.95\linewidth}
\begin{figure}
    \centering
    \begin{tabular}{c}
    \centering
    \includegraphics[width=\swone]{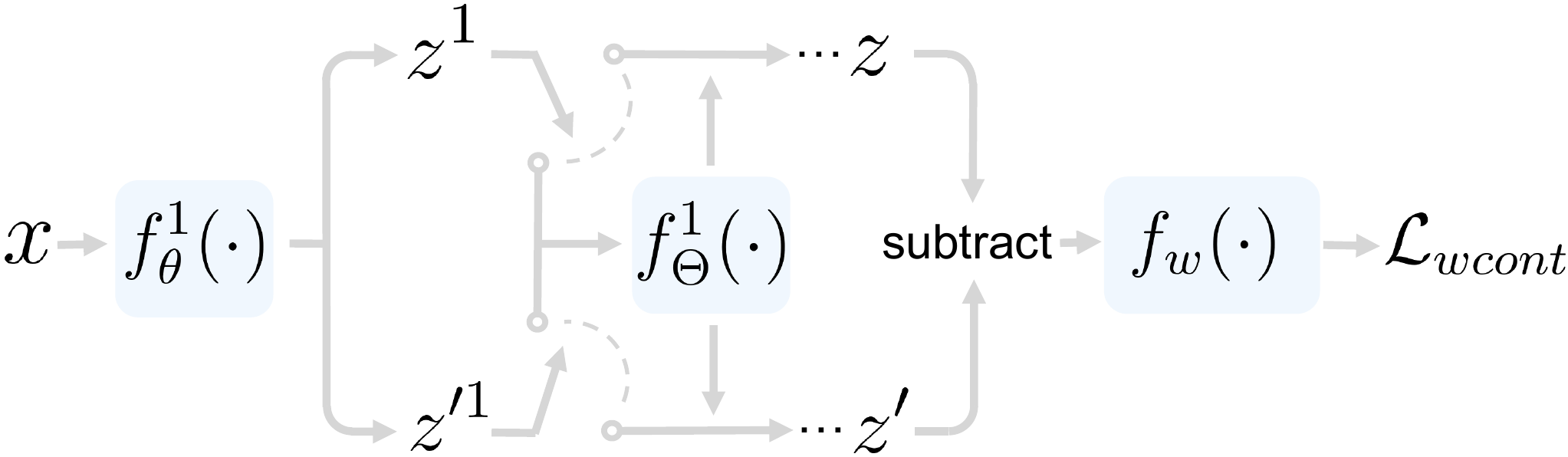}
    \end{tabular}
    \vspace{-0.3 cm}
    \caption{Test adaptation process of ITTA. Different from that in the training stage, we include additional adaptive parameters $f_{\Theta}$ after each block of the feature extractor $f_{\theta}$. For each test sample $x$, the intermediate representations $z^i$ and $z'^i$ obtained from $f_{\theta}^i$ are passed to $f_{\Theta}^i$ before going to the next block $f_{\theta}^{i+1}$. We use the learnable consistency loss $\mathcal{L}_{wcont}$ as the objective to update $f_{\Theta}$. Please refer to our text for details.}
    \label{fig testpipe}
    \vspace{-0.5 cm}
\end{figure}

Combining $\mathcal{L}_{wcont}$ with the main loss $\mathcal{L}_{main}$ which applies the cross-entropy loss (CE) for both the original and augmented inputs (\ie $\mathcal{L}_{main} = \text{CE}(f_{\phi}(z),y) + \text{CE}(f_{\phi}(z'),y)$, where $f_{\phi}$ is the classifier parameterized by $\phi$, and $y$ is the corresponding label), the objective for the feature extractor and classifier can be formulated into,
\begin{equation}
\label{eq allloss}
\min\nolimits_{\{\theta, \phi\}} \mathcal{L}_{main} + \alpha\mathcal{L}_{wcont},
\end{equation}
where $\alpha$ is the weight parameter that balances the contributions from the two terms. A simple illustration of the workflow is shown in Figure~\ref{fig pipeline}.

From Eq.~\eqref{eq allloss}, the expected gradients for the feature extractor from $\mathcal{L}_{main}$ and $\mathcal{L}_{wcont}$ can be represented as,
\begin{numcases}{}
\textbf{g}_{main} = \nabla_{\theta} (\text{CE}(f_{\phi}(z),y) + \text{CE}(f_{\phi}(z'),y)), \label{eq gmain}\\
\textbf{g}_{wcont} =\nabla_{\theta} \Vert f_w(z - z')\Vert. \label{eq gwcont}
\end{numcases}
We observe that the direction of $\textbf{g}_{wcont}$ is also determined by the weight subnetwork $f_w(\cdot)$, which should be close with $\textbf{g}_{main}$ to ensure alignment between $\mathcal{L}_{main}$ and $\mathcal{L}_{wcont}$~\cite{sun2020test,liu2021ttt++}. To this end, we propose a straightforward solution by enforcing equality between the normalized versions of $\textbf{g}_{main}$ and $\textbf{g}_{wcont}$, and we use this term as the objective for updating $f_w(\cdot)$,  which gives,
\begin{equation}
    \label{eq alignloss}
    \min_w\mathcal{L}_{align},~~~~\text{s.t.}~~\mathcal{L}_{align} = \Vert \hat{\textbf{g}}_{main} -  \hat{\textbf{g}}_{wcont} \Vert,
\end{equation}
where $\hat{\textbf{g}}_{main} = \frac{\textbf{g}_{main} - \mathbb{E}_{g_{main}}}{\sigma_{g_{main}}}$, and similar for $\hat{\textbf{g}}_{wcont}$.

In our implementation, we update $\{\theta, \phi\}$ and $w$ in an alternative manner. Pseudo code of the training process are shown in Algorithm~\ref{alg 1}.

\begin{algorithm}[t]
\caption{Pseudo code of the test phase of ITTA in a PyTorch-like style.}
\footnotesize
\comment{\# ~~ $f_{\theta}, f_{\phi}$: ~feature extractor, classifier}\\
\comment{\# ~~ $f_w, f_{\Theta}$: ~weight subnetwork, additional adaptive blocks}\\
\comment{\# ~~ $m, \textbf{0}$: ~total number of blocks in $f_{\theta}$, all zero tensor} \vspace{0.3 cm}\\
\comment{\# ~~test process}\\
for $x$ in test\_loader: ~~\comment{\# ~~load a test batch}\\
\begin{algorithmic}[]
\vspace{-0.4 cm}
\STATE def forward\_process($x$):\\
\STATE \D $z^1,z'^1 = f_{\Theta}^1.\text{forward}((f_{\theta}^1.$forward($x$))) ~~\comment{\# ~~first blocks}\\
\STATE \D for $i$ in range(2, $m+1$): ~~\comment{\# ~~the following $m-1$ blocks}
\STATE \D \D $z^i,z'^i = f_{\theta}^i.$forward($z^{i-1}$),
$f_{\theta}^i.$forward($z'^{i-1}$)\\
\STATE \D \D $z^i,z'^i = f_{\Theta}^i.$forward($z^{i}$),
$f_{\Theta}^i.$forward($z'^{i}$) \vspace{0.3CM}\\
\STATE \D return $z^i,~~z'^i$ \vspace{0.3cm}\\
\STATE \comment{\# ~~test adaptation phase: SGD update additional adaptive parameters}\\
\STATE $z,~~ z'$ = forward\_process($x$) \\
\STATE $\mathcal{L}_{wcont}=$ MSELoss($f_w.\text{forward}(z-z')$,~~\textbf{0})\\
\STATE $f_{\Theta}$.params.zero\_grad()
\STATE $\mathcal{L}_{wcont}.$backward() \\
\STATE update($f_{\Theta}.\text{params}$) \vspace{0.3 cm}\\
\STATE \comment{\# ~~final prediction}\\
\STATE $z,~~ \_$ = forward\_process($x$) \\
\STATE result = $f_{\phi}$.forward(z)
\end{algorithmic}
\label{alg 2}
\end{algorithm}

\begin{table*}[t]
\centering
\caption{Multi sources domain generalization. Experiments are conducted on the DomainBed benchmark \cite{gulrajani2020search}. All methods are examined for 60 trials in each unseen domain. Top5 accumulates the number of datasets where a method achieves the top 5 performances. The score here accumulates the numbers of the dataset where a specific art obtains larger accuracy than ERM on account of the variance. Best results are colored as \red{red}. Among the 22 methods compared, less than a quarter outperforms ERM in most datasets (Score $\geq 3$).}
\scalebox{1}{
\begin{tabular}{lC{1.7cm}C{1.7cm}C{1.7cm}C{1.7cm}C{1.7cm}C{0.9cm}| C{0.9cm} C{0.9cm}}
\toprule 
& PACS & VLCS & OfficeHome & TerraInc & DomainNet & Avg. & Top5$\uparrow$ &Score$\uparrow$\\
\hline \hline
MMD \cite{li2018domain} &81.3 $\pm$ 0.8 &74.9 $\pm$ 0.5 &59.9 $\pm$ 0.4 &42.0 $\pm$ 1.0 &7.9 $\pm$ 6.2 &53.2 &1 &2\\
RSC  \cite{huang2020self} &80.5 $\pm$ 0.2 &75.4 $\pm$ 0.3 &58.4 $\pm$ 0.6 &39.4 $\pm$ 1.3 &27.9 $\pm$ 2.0 &56.3 &0 &1\\
IRM \cite{arjovsky2019invariant} &80.9 $\pm$ 0.5 & 75.1 $\pm$ 0.1 &58.0 $\pm$ 0.1 &38.4 $\pm$ 0.9 &30.4 $\pm$ 1.0 &56.6 &0 &1\\
ARM \cite{zhang2020adaptive} &80.6 $\pm$ 0.5 &75.9 $\pm$ 0.3 &59.6 $\pm$ 0.3 &37.4 $\pm$ 1.9 &29.9 $\pm$ 0.1 &56.7 &0 &0\\
DANN \cite{ganin2016domain} &79.2 $\pm$ 0.3 &76.3 $\pm$ 0.2 &59.5 $\pm$ 0.5 &37.9 $\pm$ 0.9 &31.5 $\pm$ 0.1 &56.9 &1 &1\\
GroupGRO \cite{sagawa2019distributionally} &80.7 $\pm$ 0.4 &75.4 $\pm$ 1.0 &60.6 $\pm$ 0.3 &41.5 $\pm$ 2.0 &27.5 $\pm$ 0.1 &57.1 &0 &1\\
CDANN \cite{li2018deep} &80.3 $\pm$ 0.5 &76.0 $\pm$ 0.5 &59.3 $\pm$ 0.4 &38.6 $\pm$ 2.3 &31.8 $\pm$ 0.2 &57.2 &0 &0\\
VREx \cite{krueger2021out} &80.2 $\pm$ 0.5 &75.3 $\pm$ 0.6 &59.5 $\pm$ 0.1 &\red{43.2 $\pm$ 0.3} &28.1 $\pm$ 1.0 &57.3 &1 &1\\
CAD \cite{ruan2021optimal} &81.9 $\pm$ 0.3 &75.2 $\pm$ 0.6 &60.5 $\pm$ 0.3 &40.5 $\pm$ 0.4 &31.0 $\pm$ 0.8 &57.8 &1 &2\\
CondCAD \cite{ruan2021optimal} &80.8 $\pm$ 0.5 &76.1 $\pm$ 0.3 &61.0 $\pm$ 0.4 &39.7 $\pm$ 0.4 &31.9 $\pm$ 0.7 &57.9 &0 &1\\
MTL \cite{blanchard2017domain} &80.1 $\pm$ 0.8 &75.2 $\pm$ 0.3 &59.9 $\pm$ 0.5 &40.4 $\pm$ 1.0 &35.0 $\pm$ 0.0 &58.1 &0 &0\\
ERM \cite{vapnik1999nature} &79.8 $\pm$ 0.4 &75.8 $\pm$ 0.2 &60.6 $\pm$ 0.2 &38.8 $\pm$ 1.0 &35.3 $\pm$ 0.1 &58.1 &1 &-\\
MixStyle \cite{zhou2021domain} &82.6 $\pm$ 0.4 &75.2 $\pm$ 0.7 &59.6 $\pm$ 0.8 &40.9 $\pm$ 1.1 &33.9 $\pm$ 0.1 &58.4 &1 &1\\
MLDG \cite{li2018learning} &81.3 $\pm$ 0.2 &75.2 $\pm$ 0.3 &60.9 $\pm$ 0.2 &40.1 $\pm$ 0.9 &35.4 $\pm$ 0.0 &58.6 &1 &1\\
Mixup \cite{yan2020improve} &79.2 $\pm$ 0.9 &76.2 $\pm$ 0.3 &61.7 $\pm$ 0.5 &42.1 $\pm$ 0.7 &34.0 $\pm$ 0.0 &58.6 &2 &2\\
Fishr  \cite{rame2021ishr} &81.3 $\pm$ 0.3 &76.2 $\pm$ 0.3 &60.9 $\pm$ 0.3 &42.6 $\pm$ 1.0 &34.2 $\pm$ 0.3 &59.0 &2 &2\\
SagNet \cite{nam2021reducing} &81.7 $\pm$ 0.6 &75.4 $\pm$ 0.8 &62.5 $\pm$ 0.3 &40.6 $\pm$ 1.5 &35.3 $\pm$ 0.1 &59.1 &1 &2\\
SelfReg \cite{kim2021selfreg} &81.8 $\pm$ 0.3 &76.4 $\pm$ 0.7 &62.4 $\pm$ 0.1 &41.3 $\pm$ 0.3 &34.7 $\pm$ 0.2 &59.3 &2 &3\\
Fish \cite{shi2021gradient} &82.0 $\pm$ 0.3 &\red{76.9 $\pm$ 0.2} &62.0 $\pm$ 0.6 &40.2 $\pm$ 0.6 &35.5 $\pm$ 0.0 &59.3 &3 &4\\
CORAL \cite{sun2016deep} &81.7 $\pm$ 0.0 &75.5 $\pm$ 0.4 &62.4 $\pm$ 0.4 &41.4 $\pm$ 1.8 &36.1 $\pm$ 0.2 &59.4 &2 &3\\
SD \cite{pezeshki2021gradient} &81.9 $\pm$ 0.3 &75.5 $\pm$ 0.4 &\red{62.9 $\pm$ 0.2} &42.0 $\pm$ 1.0 &\red{36.3 $\pm$ 0.2} &59.7 &4 &4\\
Ours &\red{83.8 $\pm$ 0.3} &\red{76.9 $\pm$ 0.6} &62.0 $\pm$ 0.2 &\red{43.2 $\pm$ 0.5} &34.9 $\pm$ 0.1 &\red{60.2} &4 &4\\
\bottomrule
\end{tabular}}
\label{tab multi_results}
\vspace{-0.5 cm}
\end{table*}

\subsection{Including Additional Adaptive Parameters}
Selecting expressive and reliable parameters to update during the test phase is also essential in the TTT framework~\cite{wang2020tent}. Some strategies decide to update all the parameters from the feature extractor \cite{li2021test,bartler2022mt3}, while others use only the parameters from the specific layers for updating \cite{wang2020tent,you2021test}. Given the fact that the sizes of current deep models are often very large and still growing, exhaustively trying different combinations among the millions of candidates seems to be an everlasting job. As there are no consensuses on which parameter should be updated, we suggest another easy alternative in this work. 

Specifically, assuming there are a total of $m$ blocks in the pretrained feature extractor $f_{\theta}(\cdot)$, and the $i$-th block can be denoted as~ $f_{\theta}^i(\cdot)$. Then the intermediate representation $z^i$ from $f_{\theta}^i(\cdot)$ can be formulated as,
\begin{equation}
    \label{eq intermediate_layer}
    z^i = f_{\theta}^i(z^{i-1}),~~\text{s.t.}~~z^1 = f_{\theta}^1(x).
\end{equation}
We propose to include additional adaptive block $f_{\Theta}$ that is parameterized by $\Theta$ after each block of $f_{\theta}$ during the test-time adaptation phase, which reformulates Eq.~\eqref{eq intermediate_layer} into,
\begin{equation}
    \label{eq add_params}
    z^i = f_{\Theta}^i(f_{\theta}^i(z^{i-1})),~~\text{s.t.}~~z^1 = f_{\Theta}^1(f_{\theta}^1(x)),
\end{equation}
where $f_{\Theta}(\cdot)$ does not change the dimension and sizes of the intermediate representations. In our work, we use a structure similar to $f_w$ to implement $f_{\Theta}$. Note $z^m$ is simplified as $z$ in this phase, and the same process is applied for obtaining $z'$.

Then, in the test-time adaptation phase, we suggest only updating the new adaptive parameters via the learned consistency loss. The optimization process can be written as, 
\begin{equation}
    \label{eq test_update}
    \min_{\Theta} \Vert f_w(z-z')\Vert,~~\text{s.t.}~~\{z,z'\}=f_{\Theta}(f_{\theta}(x)).
\end{equation}
Note that different from the training phase, $x$ in this stage is from the target domain $\mathcal{D}_t$, and we use the online setting in \cite{sun2020test} for updating. A simple illustration of the test adaptation pipeline is shown in Figure~\ref{fig testpipe}.

For the final step, we use the original representation obtained from the pretrained feature extractor and the adapted adaptive parameters for prediction. Pseudo code of the test stage are shown in Algorithm~\ref{alg 2}.

\section{Experiments}
\subsection{Settings}
\label{sec settings}
\noindent\textbf{Datasets.}
We evalute ITTA on five benchmark datasets: \textbf{PACS} \cite{li2017deeper} which consists of 9,991 images from 7 categories. This dataset is probably the most widely-used DG benchmark owing to its large distributional shift across 4 domains including art painting, cartoon, photo, and sketch; \textbf{VLCS} \cite{fang2013unbiased} contains 10,729 images of 5 classes from 4 different datasets (\ie domains) including PASCAL VOC 2007 \cite{everingham2010pascal}, LabelMe \cite{russell2008labelme}, Caltech \cite{fei2004learning}, and Sun \cite{xiao2010sun} where each dataset is considered a domain in DG; \textbf{OfficeHome} \cite{venkateswara2017deep} is composed of 15,588 images from 65 classes in office and home environments, and those images can be categorized into 4 domains (\ie artistic, clipart, product, and real world); \textbf{TerraInc} \cite{beery2018recognition} has 24,788 images from 10 classes. Those images are wild animals taken from 4 different locations (\ie domains) including L100, L38, L43, and L46; \textbf{DomainNet} \cite{peng2019moment} which contains 586,575 images from 345 classes, and the images in it can be depicted in 6 styles (\ie clipart, infograph, painting, quickdraw, real, and sketch).

\begin{table*}[t]
\centering
\caption{Single source domain generalization. Experiments are conducted on the PACS dataset \cite{li2017deeper}. Here A, C, P, and S are the art, cartoon, photo, and sketch domains in PACS. A$\rightarrow$C represents models trained on the art domain and tested on the cartoon domain, and similar for others. All methods are examined for 60 trials in each unseen domain. Best results are colored as \red{red}.}
\scalebox{0.725}{
\begin{tabular}{L{1.4cm}cccccccccccc|c}
\toprule 
& A$\rightarrow$C & A$\rightarrow$P & A$\rightarrow$S & C$\rightarrow$A & C$\rightarrow$P & C$\rightarrow$S & P$\rightarrow$A & P$\rightarrow$C & P$\rightarrow$S & S$\rightarrow$A & S$\rightarrow$C & S$\rightarrow$P  &Avg.\\
\hline \hline
RSC  &66.3$\pm$1.3 &88.2$\pm$0.6 &57.2$\pm$3.1 &65.8$\pm$1.5 &82.4$\pm$0.6 &68.7$\pm$2.5 &60.5$\pm$2.0 &41.3$\pm$6.0 &53.1$\pm$2.8 &53.8$\pm$1.6 &65.9$\pm$0.7 &48.4$\pm$1.9 &62.6\\
Fish  &67.1$\pm$0.5 &89.2$\pm$1.8 &57.0$\pm$0.2 &66.7$\pm$1.0 &85.6$\pm$0.4 &64.5$\pm$3.6 &55.1$\pm$2.1 &33.9$\pm$2.3 &51.2$\pm$4.2 &59.1$\pm$3.2 &67.1$\pm$0.9 &58.4$\pm$1.2 &62.9\\
CDANN &66.5$\pm$1.7 &92.2$\pm$0.6 &65.0$\pm$0.9 &70.6$\pm$0.1 &82.9$\pm$1.4 &67.7$\pm$3.0 &60.6$\pm$0.3 &42.2$\pm$6.4 &46.9$\pm$9.9 &51.4$\pm$2.3 &60.7$\pm$1.2 &51.9$\pm$0.4 &63.2\\
SelfReg &63.9$\pm$1.9 &90.1$\pm$1.0 &56.8$\pm$2.2 &70.2$\pm$2.3 &85.4$\pm$0.3 &70.2$\pm$2.2 &60.9$\pm$2.6 &38.8$\pm$4.0 &50.5$\pm$3.2 &54.5$\pm$4.7 &66.2$\pm$1.2 &51.7$\pm$4.1 &63.3\\
DANN &67.5$\pm$1.6 &91.2$\pm$1.3 &67.5$\pm$1.3 &70.6$\pm$1.0 &81.4$\pm$0.4 &66.6$\pm$1.1 &54.1$\pm$2.3 &33.5$\pm$2.7 &52.8$\pm$2.3 &53.8$\pm$1.7 &64.4$\pm$0.7 &58.9$\pm$0.8 &63.5\\
CAD  &67.1$\pm$1.5 &89.6$\pm$0.4 &60.2$\pm$0.2 &67.7$\pm$3.1 &83.7$\pm$1.4 &70.2$\pm$2.6 &60.6$\pm$2.6 &38.3$\pm$3.7 &53.8$\pm$3.2 &50.7$\pm$1.6 &65.8$\pm$1.3 &54.4$\pm$1.7 &63.5\\
GroupGRO  &66.5$\pm$1.2 &90.5$\pm$1.5 &58.9$\pm$2.5 &70.8$\pm$0.9 &85.7$\pm$1.2 &69.7$\pm$1.8 &62.3$\pm$2.1 &41.1$\pm$2.7 &48.2$\pm$4.1 &54.8$\pm$0.5 &65.2$\pm$1.6 &53.9$\pm$1.4 &64.0\\
MTL &67.3$\pm$1.0 &90.1$\pm$1.0 &58.9$\pm$0.7 &70.2$\pm$1.8 &84.2$\pm$2.2 &71.9$\pm$0.7 &58.3$\pm$2.7 &38.5$\pm$2.7 &52.8$\pm$1.5 &55.4$\pm$3.1 &66.1$\pm$1.3 &55.2$\pm$2.6 &64.1\\
IRM  &67.5$\pm$1.8 &93.0$\pm$0.5 &62.9$\pm$4.7 &67.6$\pm$1.3 &83.8$\pm$0.4 &68.9$\pm$0.8 &63.7$\pm$1.8 &39.9$\pm$3.7 &49.0$\pm$5.4 &54.9$\pm$1.4 &63.1$\pm$2.1 &54.9$\pm$1.4 &64.1\\
ARM &66.0$\pm$2.4 &91.2$\pm$0.7 &58.7$\pm$6.9 &70.6$\pm$0.8 &84.2$\pm$1.0 &69.1$\pm$0.9 &59.2$\pm$1.8 &42.1$\pm$5.6 &52.1$\pm$3.0 &60.0$\pm$0.6 &62.9$\pm$3.3 &53.8$\pm$2.0 &64.2\\
Mixup &65.5$\pm$0.8 &87.8$\pm$0.3 &57.2$\pm$1.0 &71.4$\pm$1.1 &83.1$\pm$1.8 &68.0$\pm$3.0 &59.6$\pm$1.7 &37.2$\pm$2.7 &56.5$\pm$3.8 &55.0$\pm$2.2 &66.2$\pm$1.5 &\red{62.7$\pm$4.2} &64.2\\
CORAL &66.8$\pm$0.5 &90.3$\pm$0.7 &61.5$\pm$1.9 &67.9$\pm$2.1 &85.4$\pm$0.3 &70.4$\pm$1.3 &55.9$\pm$2.9 &40.4$\pm$4.9 &49.8$\pm$8.5 &55.8$\pm$2.1 &67.6$\pm$0.9 &58.9$\pm$3.8 &64.2\\
SD &67.1$\pm$1.3 &91.7$\pm$1.2 &63.7$\pm$4.1 &70.3$\pm$0.9 &84.4$\pm$0.7 &69.4$\pm$2.3 &57.5$\pm$2.5 &42.6$\pm$0.8 &47.7$\pm$1.7 &55.9$\pm$2.4 &65.7$\pm$0.8 &55.8$\pm$2.1 &64.3\\
MMD &67.1$\pm$1.4 &88.0$\pm$0.8 &63.6$\pm$1.6 &70.0$\pm$1.1 &83.6$\pm$0.2 &70.2$\pm$1.0 &58.8$\pm$2.6 &40.3$\pm$1.0 &52.3$\pm$2.4 &57.4$\pm$1.9 &\red{68.7$\pm$0.9} &52.7$\pm$3.7 &64.4\\
MLDG  &67.3$\pm$2.0 &90.8$\pm$0.5 &64.4$\pm$0.9 &70.8$\pm$1.0 &84.2$\pm$0.3 &69.7$\pm$1.8 &61.6$\pm$1.0 &41.3$\pm$5.1 &50.4$\pm$0.2 &49.9$\pm$2.5 &66.8$\pm$0.4 &58.7$\pm$3.4 &64.7\\
CondCAD  &66.9$\pm$1.4 &92.3$\pm$0.7 &60.8$\pm$4.5 &71.0$\pm$0.6 &84.7$\pm$1.1 &\red{72.6$\pm$0.5} &61.2$\pm$1.5 &40.7$\pm$3.6 &55.7$\pm$1.6 &52.3$\pm$1.7 &64.2$\pm$0.4 &55.3$\pm$1.2 &64.8\\
ERM  &67.3$\pm$0.7 &91.7$\pm$0.9 &60.1$\pm$4.7 &70.4$\pm$0.6 &82.3$\pm$2.7 &68.1$\pm$0.9 &59.6$\pm$1.8 &44.7$\pm$2.8 &56.5$\pm$2.7 &52.8$\pm$2.3 &68.1$\pm$0.7 &58.4$\pm$0.9 &65.0\\
VREx &67.1$\pm$1.5 &91.0$\pm$1.0 &62.6$\pm$3.5 &71.1$\pm$2.4 &84.1$\pm$0.9 &71.7$\pm$1.3 &62.4$\pm$3.1 &37.7$\pm$3.3 &53.6$\pm$2.3 &\red{60.6$\pm$1.6} &66.7$\pm$0.8 &57.5$\pm$1.4 &65.5\\
Fishr &67.9$\pm$1.9 &\red{92.7$\pm$0.3} &62.4$\pm$4.7 &71.2$\pm$0.5 &83.4$\pm$0.6 &70.2$\pm$1.1 &60.0$\pm$2.3 &42.7$\pm$3.2 &57.1$\pm$3.9 &55.7$\pm$3.7 &68.4$\pm$1.0 &62.0$\pm$3.1 &66.1\\
SagNet &67.6$\pm$1.4 &92.3$\pm$0.5 &59.5$\pm$1.7 &71.8$\pm$0.3 &82.8$\pm$0.6 &69.9$\pm$1.8 &62.5$\pm$2.5 &45.2$\pm$2.5 &\red{64.1$\pm$2.0} &55.8$\pm$1.1 &65.7$\pm$1.4 &55.9$\pm$3.5 &66.1\\
MixStyle  &68.5$\pm$2.0 &91.2$\pm$1.6 &\red{65.1$\pm$0.7} &73.2$\pm$1.3 &85.0$\pm$0.8 &71.7$\pm$1.5 &63.6$\pm$1.7 &46.3$\pm$1.1 &51.6$\pm$3.7 &54.2$\pm$1.5 &67.0$\pm$3.4 &58.3$\pm$1.4 &66.3\\
Ours &\red{68.9$\pm$0.6} &92.4$\pm$0.1 &62.5$\pm$0.6 &\red{75.3$\pm$0.4} &\red{85.9$\pm$0.3} &70.2$\pm$1.4 &\red{66.5$\pm$1.1} &\red{52.2$\pm$2.7} &63.8$\pm$1.1 &57.6$\pm$3.7 &68.0$\pm$1.3 &57.9$\pm$2.0 &\red{68.4}\\
\bottomrule
\end{tabular}}
\label{tab single_results}
\vspace{-0.5 cm}
\end{table*}

\noindent\textbf{Implementation details.}
For all the experiments, we use the ImageNet \cite{deng2009imagenet} pretrained ResNet18 \cite{he2016deep} backbone that with 4 blocks as the feature extractor $f_{\theta}$, which could enlarge the gaps in DG compared to larger models \cite{ye2022ood}. Correspondingly, we also include 4 blocks of additional adaptive parameters (\ie $f_{\Theta}$), and each block is implemented with 5 layers of learnable parameters with weight initialized as all ones and bias initialized as all zeros. For the weight subnetwork $f_w$, we use 10 layers of learnable parameters with the initialization skill similar to that of $f_{\Theta}$. The classifier $f_{\phi}$ is an  MLP layer provided by the Domainbed benchmark \cite{gulrajani2020search}. For the weight parameter $\alpha$ in Eq.~\eqref{eq allloss}, we set it to be 1 for all experiments (please refer to our supplementary material for analysis). The random seeds, learning rates, batch size, and augmentation skills are all dynamically set for all the compared arts according to \cite{gulrajani2020search}.

\noindent\textbf{Training and evaluation details.}
For all the compared methods, we conduct 60 trials on each source domain, and each with 5,000 iteration steps. During the training stage, we split the examples from training domains to 8:2 (train:val) where the training and validation samples are dynamically selected among different training trials. During test, we select the model that performs the best in the validation samples and test it on the target domains. The strategy is referred to as the ``training-domain validate set” model selection method in \cite{gulrajani2020search}. For each domain in different datasets, the final performance is the average accuracy from the 60 trials. 

\subsection{Multi-Source Generalization}
In these experiments, all five benchmark datasets aforementioned are used for evaluation, and the leave-one-out strategy is adopted for training (\ie with $S=\vert \mathcal{D}_s \cup \mathcal{D}_t\vert \footnote{We use $\vert \cdot \vert$ to denote the number of domains in the environment.} - 1$, and $T=1$). Results are shown in Table~\ref{tab multi_results}. We note that ERM method obtains favorable performance against existing arts. In fact, as a strong baseline, ERM is superior to half of the methods in the term of average accuracy, and only 5 arts (\ie SelfReg \cite{kim2021selfreg}, Fish \cite{shi2021gradient}, CORAL \cite{sun2016deep}, SD \cite{pezeshki2021gradient}, and ours) among the compared 22 methods outperforms ERM in most datasets (\ie with Score~$\geq$~3). In comparison, the proposed ITTA is more effective than all other models on average. In particular, ITTA achieves the best performances in 3 out of the 5 benchmarks (\ie PACS, VLCS, and TerraInc datasets) and 4 in the top 5. Note that although our method does not obtain the best performances in the OfficeHome and DomainNet benchmarks, it still outperforms more than half of the existing models. The results validate the effectiveness of our method when tested in the multi-source setting. We present results of average accuracy in each domain from different datasets in the supplementary material. Please refer to it for details.

\subsection{Single-Source Generalization}
In these experiments, we adopt the widely-used PACS~\cite{li2017deeper} benchmark for evaluation, and the models are trained on one domain while tested on the remaining three (\ie with $S=1$, and $T=3$). Although some approaches, such as MLDG \cite{li2018learning} and Fishr \cite{rame2021ishr}, may require more than one domain information for their trainings, we can simulate multi-domain information using only the source domain, and thus the experimental settings are still feasible for them. Compared to the multi-source generalization task, the single-source generalization is considered more difficult due to the limited domain information during the training phase. 
Evaluation results are presented in Table~\ref{tab single_results}. We note that the ERM method outperforms most state-of-the-art models, and only 5 models, including VREx \cite{krueger2021out}, Fishr  \cite{rame2021ishr}, SagNet \cite{nam2021reducing}, MixStyle \cite{zhou2021domain}, and the proposed ITTA, can obtain better results than ERM in the term of average accuracy. Meanwhile, our method achieves the best performances when trained in 5 out of the 12 source domain, and it obtains the best performance on average, leading more than $2\%$ than the second best (\ie MixStyle \cite{zhou2021domain}) and $3\%$ the ERM method.

In line with the findings in \cite{gulrajani2020search}, we notice that the naive ERM method \cite{vapnik1999nature} can indeed perform favorably against most existing models under rigorous evaluation protocol. As a matter of fact, the proposed method is the only one that consistently outperforms ERM in both the multi-source and single-source settings. These results indicate that DG remains challenging for current efforts that aim to ease the distribution shift only through training data, and using the proposed improved TTT strategy may be a promising direction for solving DG.

\begin{table}[t]
\centering
\caption{Evaluations of different TTT-based models in the unseen domain from PACS \cite{li2017deeper}. The reported accuracies ($\%$) and standard deviations are computed from 60 trials in each target domain.}
\scalebox{0.77}{
\begin{tabular}{lccccc}
\toprule 
 \multirow{2}*{Model} & \multicolumn{4}{c}{Target domain} &\multirow{2}*{Avg.}\\
\cline{2-5}
 & Art & Cartoon & Photo & Sketch\\
\hline \hline
Baseline &79.9$\pm$0.5 &75.4$\pm$1.1 &94.4$\pm$0.5 &75.8$\pm$1.2 &81.4$\pm$0.5\\
TTT \cite{sun2020test} &81.5$\pm$0.8 &77.6$\pm$0.6 &94.3$\pm$0.2 &78.4$\pm$0.7 &83.0$\pm$0.2 \\
MT3 \cite{bartler2022mt3} &82.0$\pm$1.0 &76.5$\pm$1.0 &94.1$\pm$0.2 &77.7$\pm$1.3 &82.6$\pm$0.6 \\
TENT \cite{wang2020tent} &80.2$\pm$0.9 &77.2$\pm$0.8 &94.4$\pm$0.2 &77.4$\pm$0.1 &82.3$\pm$0.5 \\
Ours &84.7$\pm$0.4 &78.0$\pm$0.4 &94.5$\pm$0.4 &78.2$\pm$0.3 &83.8$\pm$0.3 \\
\bottomrule
\end{tabular}}
\label{tab allttt}
\vspace{-0.5 cm}
\end{table}

\begin{table*}[t]
\centering
\caption{Comparison between different TTT tasks and parameter selecting strategies in the unseen domain from the PACS benchmark \cite{li2017deeper}. Here the ``Ent.'', ``Rot.'', and ``$\mathcal{L}_{wcont}$'' denotes the entropy minimization task in \cite{wang2020tent}, the rotation estimation task in \cite{sun2020test}, and the proposed learnable consistency objective, the ``All", ``BN", and ``Ada." are the strategies that update all the parameters, parameters from the batch normalization layer, and the proposed strategy that updates only the new additional adaptive parameters. The reported accuracies ($\%$) and standard deviations are computed from 60 trials in each target domain.}
\scalebox{0.94}{
\begin{tabular}{l@{\extracolsep{4pt}}ccc|ccc|ccccc@{}}
\toprule 
 \multirow{2}*{Model} & \multicolumn{3}{c|}{TTT tasks} & \multicolumn{3}{c|}{Param selectings} & \multicolumn{4}{c}{Target domain} &\multirow{2}*{Avg.}\\
 \cline{2-4} \cline{5-7} \cline{8-11}
& Ent. & Rot. & $\mathcal{L}_{wcont}$ & All & BN & Ada. & Art & Cartoon & Photo & Sketch\\
\hline \hline
Ours & $-$ & $-$ & $\checkmark$ & $-$ & $-$ & $\checkmark$ &84.7$\pm$0.4 &78.0$\pm$0.4 &94.5$\pm$0.4 &78.2$\pm$0.3 &83.8$\pm$0.3 \\
\hline
Ours w/o $f_w$ & $-$ & $-$ & $-$ & $-$ & $-$ &$\checkmark$ &83.1$\pm$0.4 &74.6$\pm$0.6  &94.0$\pm$0.5 &78.0$\pm$0.8 &82.5$\pm$0.1\\
Ours w/ Ent. & $\checkmark$ & $-$ & $-$ & $-$ & $-$ &$\checkmark$ &79.9$\pm$2.4 &77.3$\pm$0.3  &94.8$\pm$0.8 &77.6$\pm$0.4 &82.4$\pm$0.8\\
Ours w/ Rot. & $-$ & $\checkmark$ & $-$ & $-$ & $-$ &$\checkmark$ &81.1$\pm$1.0 &75.2$\pm$0.5  &94.9$\pm$0.3 &77.3$\pm$0.6 &82.1$\pm$0.3\\
\hline
Ours w/o TTT & $-$ & $-$ & $\checkmark$ & $-$ & $-$ &$-$ &83.3$\pm$0.5   &76.0$\pm$0.5 &94.4$\pm$0.5  &76.7$\pm$1.4 &82.8$\pm$0.3\\
Ours w/ All  & $-$ & $-$ & $\checkmark$ & $\checkmark$ & $-$ &$-$ &83.0$\pm$0.7 &77.0$\pm$1.4 &94.5$\pm$0.7 &77.4$\pm$0.9 &83.0$\pm$0.2\\
Ours w/ BN  & $-$ & $-$ & $\checkmark$ & $-$ & $\checkmark$ &$-$ &81.8$\pm$0.5 &75.6$\pm$0.3  &94.4$\pm$0.3 &77.9$\pm$1.1 &82.4$\pm$0.5\\
\bottomrule
\end{tabular}}
\label{tab effectiveness}
\vspace{-0.3 cm}
\end{table*}

\section{Analysis}
All experiments in this section are conducted on the widely-used PACS benchmark \cite{li2017deeper} with the leave-one-out strategy. The experimental settings are the same as that illustrated in Sec. \ref{sec settings}. Please refer to our supplementary material for more analysis. 

\subsection{Compared with Other TTT-Based Models}
Using test-time adaptation to ease the distribution shift problem has been explored in previous works, such as the original TTT method \cite{sun2020test} and MT3 \cite{bartler2022mt3}. Their differences lie in that TTT uses a rotation estimation task for the test-time objective, and MT3 adopts a contrastive loss for the task and implements the overall framework using MAML \cite{finn2017model}. There is also a recently proposed TENT \cite{wang2020tent} that aims to minimize the entropy of the final results by tuning the parameters from the batch normalization (BN) layers.
To analyze the overall effectiveness of our method, we compare ITTA with these arts using the same baseline (\ie ResNet18~\cite{he2016deep} backbone with the existing augmentation skill~\cite{zhou2021domain}).

\def\swthree{0.243\linewidth}
\renewcommand{\tabcolsep}{1pt}
\begin{figure}[t]
\centering
    \begin{tabular}{cccc}
        \includegraphics[width=\swthree]{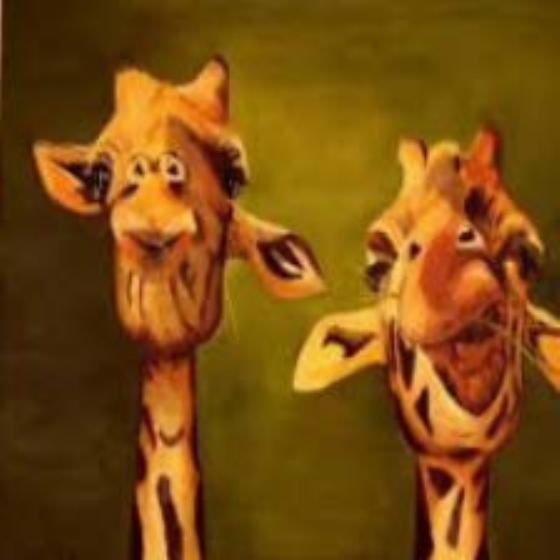}&
        \includegraphics[width=\swthree]{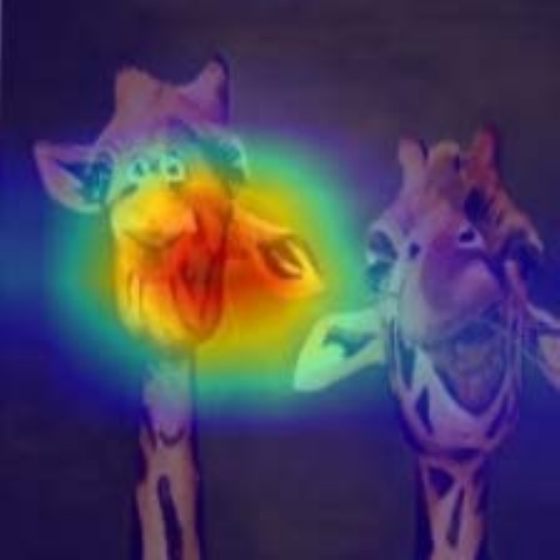}&
        \includegraphics[width=\swthree]{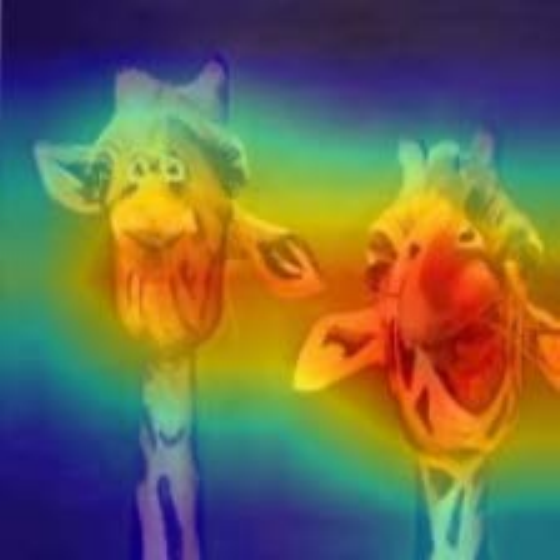}&
        \includegraphics[width=\swthree]{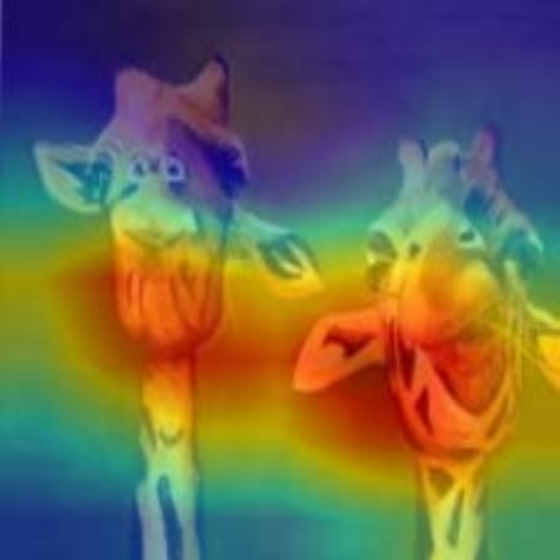}\\
        \includegraphics[width=\swthree]{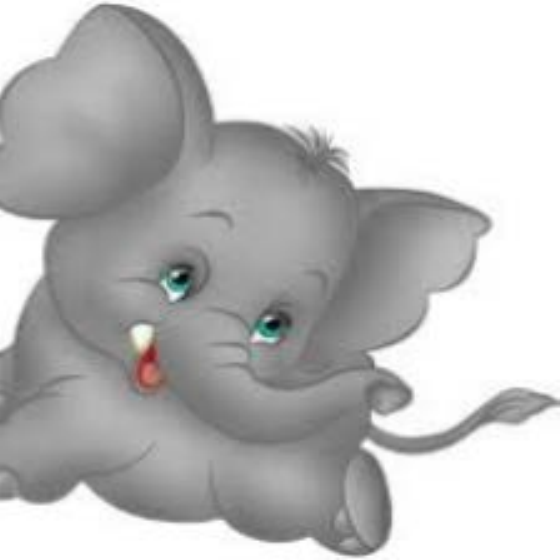}&
        \includegraphics[width=\swthree]{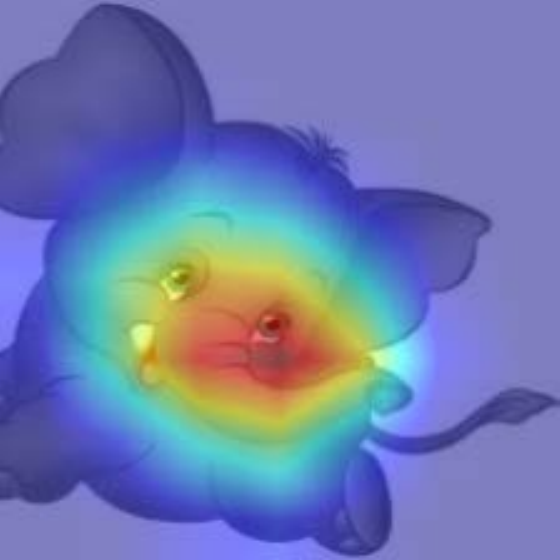}&
        \includegraphics[width=\swthree]{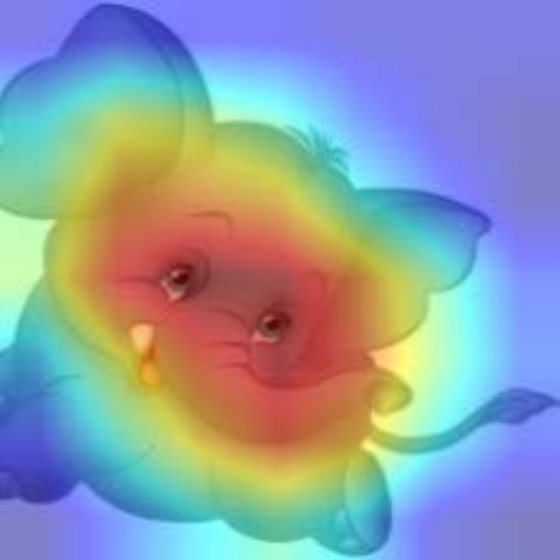}&
        \includegraphics[width=\swthree]{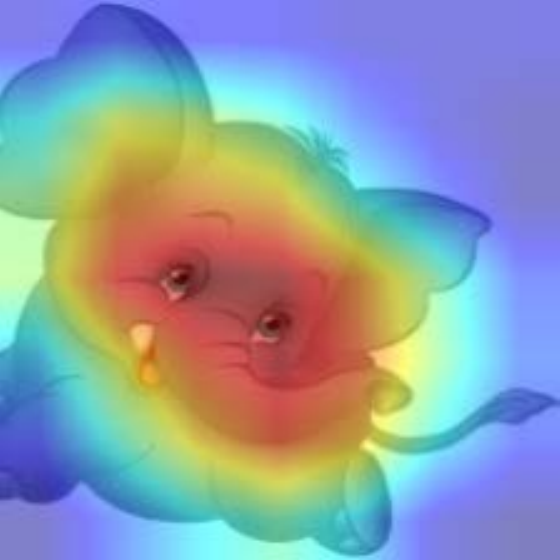}\\
        \includegraphics[width=\swthree]{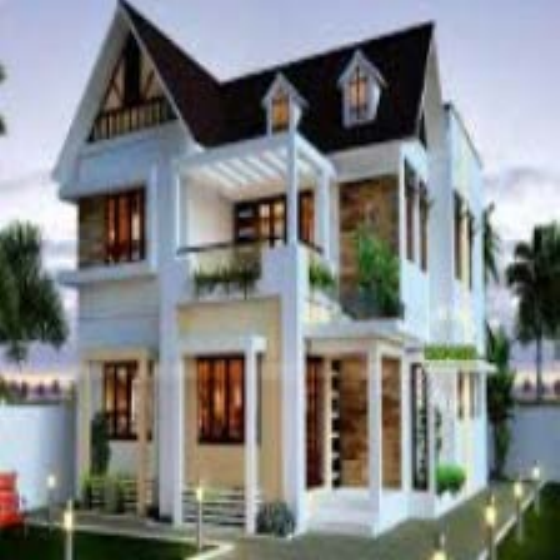}&
        \includegraphics[width=\swthree]{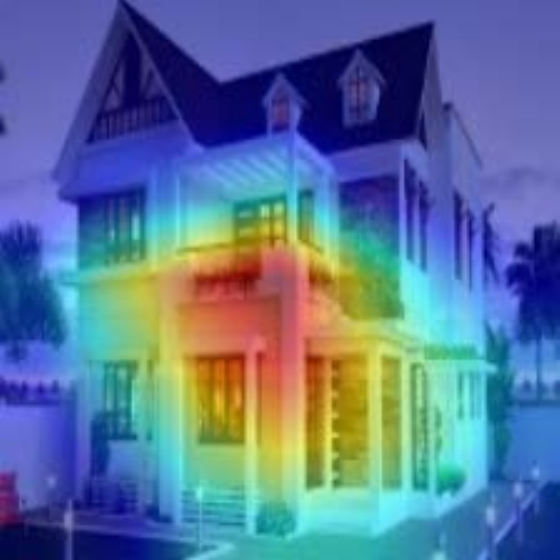}&
        \includegraphics[width=\swthree]{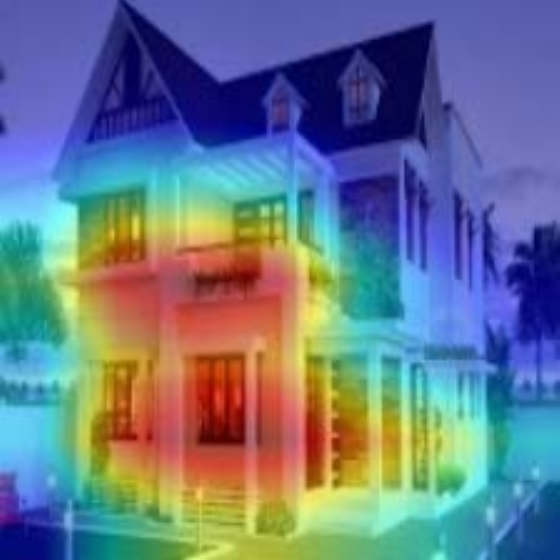}&
        \includegraphics[width=\swthree]{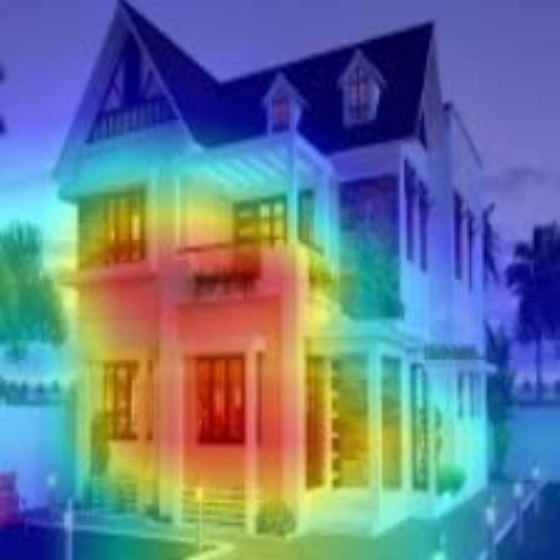}\\
        \includegraphics[width=\swthree]{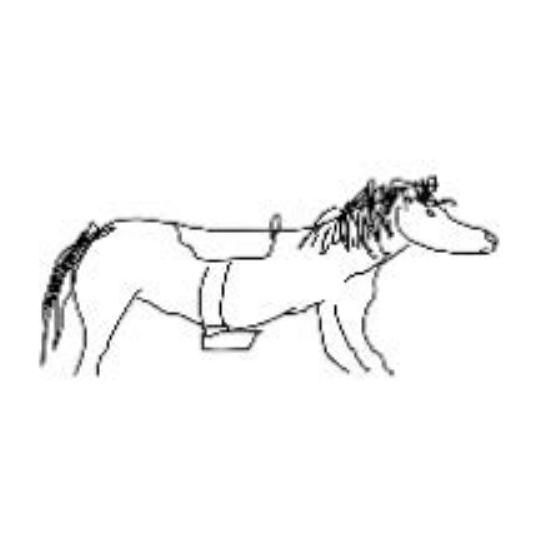}&
        \includegraphics[width=\swthree]{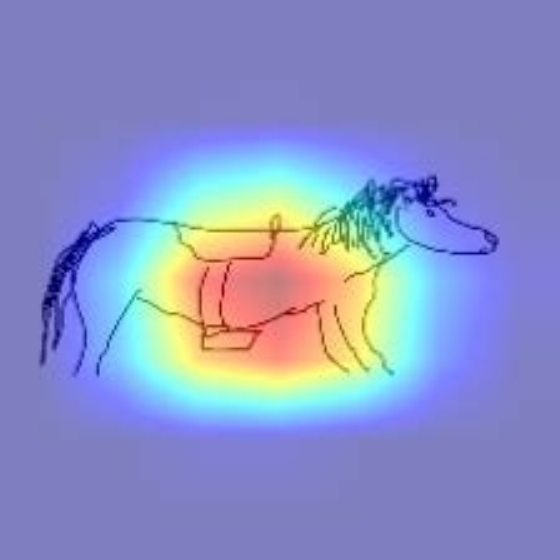}&
        \includegraphics[width=\swthree]{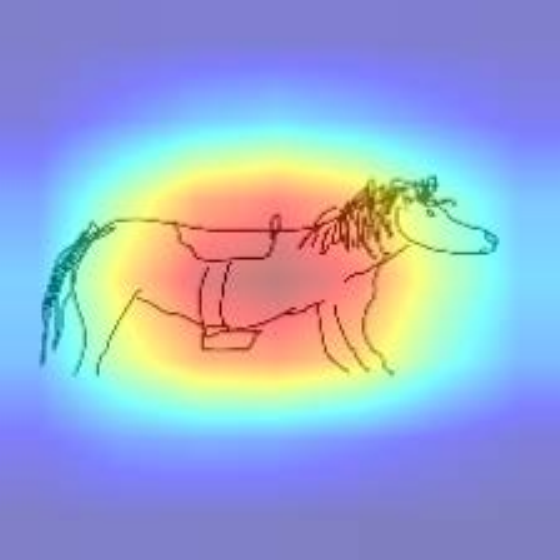}&
        \includegraphics[width=\swthree]{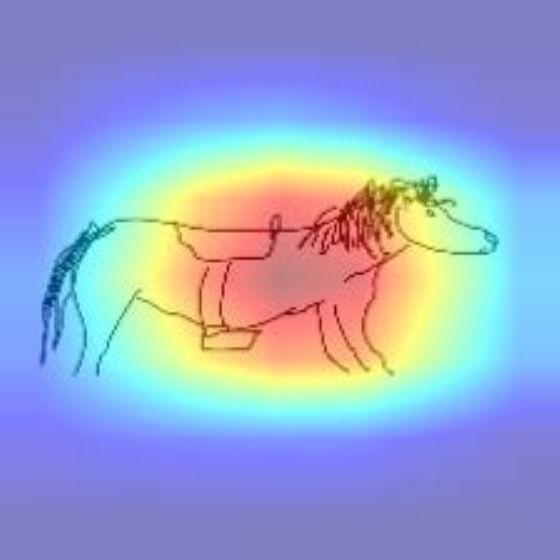}\\
        \footnotesize{\text{(a) Input}} & \footnotesize{\text{(b) Ours w/o $f_w$}}& \footnotesize{\text{(c) Ours}} & \footnotesize{\text{(d) Main}}\\
    \end{tabular}
	\caption{Grad-CAM \cite{selvaraju2017grad} visualizations from different loss terms. We use images with varying class labels from the four target domains of PACS \cite{li2017deeper} as inputs (\ie art, cartoon, photo, and sketch domains from top to bottom). Ours w/o $f_w$ is the naive consistency loss with $f_w$ disabled in Eq.~\eqref{eq wcloss}. The proposed learnable consistency loss can align well with the main classification task.}
	\label{fig cam}
	\vspace{-0.5 cm}
\end{figure}

Results are shown in Table~\ref{tab allttt}. We observe that all the compared TTT-based methods can improve the baseline model in almost all target domains except for the ``Photo" domain, which might be due to the ImageNet pretraining \cite{xu2020robust}. This phenomenon demonstrates that the TTT strategy may be a promising effort for easing the distribution shift problem. 
Meanwhile, we observe that the proposed ITTA is superior to all other approaches in most target domains and leads in the term of average accuracy. The main reason is that compared to the empirically designed TTT tasks adopted in previous works, the proposed learnable consistency loss is enforced to be more aligned with the main loss, thus more suitable for the test-time adaptation task \cite{liu2021ttt++}. Meanwhile, compared to the strategies that update the original parameters from the trained model, the adaptation of the newly included parameters is also more effective for the overall TTT framework. In the following, we provide more analysis to support these claims.

\subsection{Effectiveness of the Learnable Consistency Loss}
\label{sec loss}
To examine the effectiveness of our learnable consistency loss, we conduct ablation studies by comparing our method with the following variants. (1) Ours w/o $f_w$: we disable $f_w$ when computing the learnable consistency loss in Eq.~\eqref{eq wcloss}, which uses the naive consistency loss for the auxiliary TTT task. (2) Ours w/ Ent.: after training the model using the baseline settings (\ie ResNet18 with the augmentation strategy \cite{zhou2021domain}), we use the entropy minimization task in \cite{wang2020tent} for the TTT task. (3) Ours w/ Rot.: we use the rotation estimation task in \cite{sun2020test} for the TTT task. To ensure fair comparisons, we use the same baseline settings and include the same additional adaptive parameters for all the variants.  

Results are shown in the 4th to 6th rows Table~\ref{tab effectiveness}. We find that the results from the naive consistency loss (\ie Ours w/o $f_w$) are slightly better than that from the other two specially-designed objectives (\ie Ours w/ Ent. and Ours w/ Rot.) on average. Besides the possibility of deteriorating the performance \cite{liu2021ttt++}, our results indicate that empirically selecting a TTT task may also be far from optimal. Meanwhile, we observe that when enabling $f_w$, the proposed learnable consistency loss is superior to that without $f_w$ in all target domains, and it leads in the term of average accuracy among the variants compared, illustrating its advantage against other adopted TTT tasks. These results are not surprising. By comparing the Grad-CAM \cite{selvaraju2017grad} visualizations from the main classification task with the learnable and naive consistency losses in Figure~\ref{fig cam}, we find that the proposed learnable objective can well align with the main loss when $f_w$ is enabled as the hot zones activated by these two tasks are similar, which guarantees the improvement for the test-time adaptation \cite{liu2021ttt++,sun2020test}. Please refer to our supplementary material for more visualizations.

\subsection{Effectiveness of the Adaptive Parameters}
\label{sec param}
We compare ITTA with three variants to demonstrate the effectiveness of the proposed additional adaptive parameters. (1) Ours w/o TTT: we do not update any parameters during the test phase. This variant is used to verify whether TTT can improve the pretrained model. (2) Ours w/ ALL: similar to the updating strategy in the original TTT method \cite{sun2020test}, we update all the parameters from the feature extractor during the test phase. (3) Ours w/ BN: following the suggestion from TENT \cite{wang2020tent}, only parameters from the BN layers of the feature extractor are updated. Note the same pretrained model is shared for all variants in these experiments, and the objectives during the test adaptation phase are to minimize the same learned consistency loss.

We list the results in the last three rows in Table~\ref{tab effectiveness}. We observe that when only updating parameters from the BN layers, the performance is inferior to the strategy without test-time adaptation, and updating all the parameters does not ensure improvements in all target domains. The observations are in line with the findings in \cite{wang2020tent} that selecting reliable parameters to update is essential in the TTT system and may also interact with the choice of the TTT task. In comparison, when including additional adaptive parameters for updating, the pretrained model can be boosted in all environments. The results validate that our adaptive parameters are more effective than that selected with existing strategies \cite{sun2020test,wang2020tent} when applied with the proposed learnable test-time objective.

\subsection{Limitation}
Although the proposed learned loss can bring satisfaction improvements, we are aware that the lunch is not free. When the weight subnetwork $f_w$ is disabled, updating the joint loss in Eq.~\eqref{eq allloss} only costs 1 forward and 1 backward. However, in order to update $f_w$, we have to compute the second-order derivative in Eq.~\eqref{eq alignloss}, which will require 1 more forward and 3 more backward processes, bringing extra burden to the system. Our future efforts aim to simplify the overall optimization process and reduce the cost for ITTA.

\section{Conclusion}
In this paper, we aim to improve the current TTT strategy for alleviating the distribution shift problem in DG. First, given that the auxiliary TTT task plays a vital role in the overall framework, and an empirically selecting one that does not align with the main task may potentially deteriorate instead of improving the performance, we propose a learnable consistency loss that can be enforced to be more aligned with the main loss by adjusting its learnable parameters. This strategy is ensured to improve the model and shows favorable performance against some specially-designed objectives. Second, considering that selecting reliable and effective parameters to update during the test phase is also essential while exhaustively trying different combinations may require tremendous effort, we propose a new alternative by including new additional adaptive parameters for adaptation during the test phase. This alternative is shown to outperform some previous parameter selecting strategies via our experimental findings.
By conducting extensive experiments under a rigorous evaluation protocol, we show that our method can achieve superior performance against existing arts in both the multi-source and single-source DG tasks.

\noindent\textbf{Acknowledgements.} Liang Chen is supported by the China Scholarship Council (CSC Student ID 202008440331).

{\small
\bibliographystyle{ieee_fullname}
\bibliography{dg}

\begin{thebibliography}{10}\itemsep=-1pt

\bibitem{arjovsky2019invariant}
Martin Arjovsky, L{\'e}on Bottou, Ishaan Gulrajani, and David Lopez-Paz.
\newblock Invariant risk minimization.
\newblock {\em arXiv preprint arXiv:1907.02893}, 2019.

\bibitem{balaji2018metareg}
Yogesh Balaji, Swami Sankaranarayanan, and Rama Chellappa.
\newblock Metareg: Towards domain generalization using meta-regularization.
\newblock In {\em NeurIPS}, 2018.

\bibitem{bartler2022mt3}
Alexander Bartler, Andre B{\"u}hler, Felix Wiewel, Mario D{\"o}bler, and Bin
  Yang.
\newblock Mt3: Meta test-time training for self-supervised test-time adaption.
\newblock In {\em AISTATS}, 2022.

\bibitem{beery2018recognition}
Sara Beery, Grant Van~Horn, and Pietro Perona.
\newblock Recognition in terra incognita.
\newblock In {\em ECCV}, 2018.

\bibitem{ben2006analysis}
Shai Ben-David, John Blitzer, Koby Crammer, and Fernando Pereira.
\newblock Analysis of representations for domain adaptation.
\newblock In {\em NeurIPS}, 2006.

\bibitem{blanchard2017domain}
Gilles Blanchard, Aniket~Anand Deshmukh, Urun Dogan, Gyemin Lee, and Clayton
  Scott.
\newblock Domain generalization by marginal transfer learning.
\newblock {\em arXiv preprint arXiv:1711.07910}, 2017.

\bibitem{blanchard2011generalizing}
Gilles Blanchard, Gyemin Lee, and Clayton Scott.
\newblock Generalizing from several related classification tasks to a new
  unlabeled sample.
\newblock In {\em NeurIPS}, 2011.

\bibitem{chen2022comen}
Chaoqi Chen, Jiongcheng Li, Xiaoguang Han, Xiaoqing Liu, and Yizhou Yu.
\newblock Compound domain generalization via meta-knowledge encoding.
\newblock In {\em CVPR}, 2022.

\bibitem{chen2022mix}
Chaoqi Chen, Luyao Tang, Feng Liu, Gangming Zhao, Yue Huang, and Yizhou Yu.
\newblock Mix and reason: Reasoning over semantic topology with data mixing for
  domain generalization.
\newblock In {\em NeurIPS}, 2022.

\bibitem{chen2022contrastive}
Dian Chen, Dequan Wang, Trevor Darrell, and Sayna Ebrahimi.
\newblock Contrastive test-time adaptation.
\newblock In {\em CVPR}, 2022.

\bibitem{chen2022self}
Liang Chen, Yong Zhang, Yibing Song, Lingqiao Liu, and Jue Wang.
\newblock Self-supervised learning of adversarial example: Towards good
  generalizations for deepfake detection.
\newblock In {\em CVPR}, 2022.

\bibitem{chen2022ost}
Liang Chen, Yong Zhang, Yibing Song, Jue Wang, and Lingqiao Liu.
\newblock Ost: Improving generalization of deepfake detection via one-shot
  test-time training.
\newblock In {\em NeurIPS}, 2022.

\bibitem{chen2020simple}
Ting Chen, Simon Kornblith, Mohammad Norouzi, and Geoffrey Hinton.
\newblock A simple framework for contrastive learning of visual
  representations.
\newblock In {\em ICML}, 2020.

\bibitem{choi2022improving}
Sungha Choi, Seunghan Yang, Seokeon Choi, and Sungrack Yun.
\newblock Improving test-time adaptation via shift-agnostic weight
  regularization and nearest source prototypes.
\newblock In {\em ECCV}, 2022.

\bibitem{deng2009imagenet}
Jia Deng, Wei Dong, Richard Socher, Li-Jia Li, Kai Li, and Li Fei-Fei.
\newblock Imagenet: A large-scale hierarchical image database.
\newblock In {\em CVPR}, 2009.

\bibitem{dou2019domain}
Qi Dou, Daniel Coelho~de Castro, Konstantinos Kamnitsas, and Ben Glocker.
\newblock Domain generalization via model-agnostic learning of semantic
  features.
\newblock In {\em NeurIPS}, 2019.

\bibitem{everingham2010pascal}
Mark Everingham, Luc Van~Gool, Christopher~KI Williams, John Winn, and Andrew
  Zisserman.
\newblock The pascal visual object classes (voc) challenge.
\newblock {\em IJCV}, 88(2):303--338, 2010.

\bibitem{fang2013unbiased}
Chen Fang, Ye Xu, and Daniel~N Rockmore.
\newblock Unbiased metric learning: On the utilization of multiple datasets and
  web images for softening bias.
\newblock In {\em ICCV}, 2013.

\bibitem{fei2004learning}
Li Fei-Fei, Rob Fergus, and Pietro Perona.
\newblock Learning generative visual models from few training examples: An
  incremental bayesian approach tested on 101 object categories.
\newblock In {\em CVPR worksho}, 2004.

\bibitem{finn2017model}
Chelsea Finn, Pieter Abbeel, and Sergey Levine.
\newblock Model-agnostic meta-learning for fast adaptation of deep networks.
\newblock In {\em ICML}, 2017.

\bibitem{fleuret2021uncertainty}
Francois Fleuret et~al.
\newblock Uncertainty reduction for model adaptation in semantic segmentation.
\newblock In {\em CVPR}, 2021.

\bibitem{gandelsman2022test}
Yossi Gandelsman, Yu Sun, Xinlei Chen, and Alexei~A Efros.
\newblock Test-time training with masked autoencoders.
\newblock In {\em NeurIPS}, 2022.

\bibitem{ganin2016domain}
Yaroslav Ganin, Evgeniya Ustinova, Hana Ajakan, Pascal Germain, Hugo
  Larochelle, Fran{\c{c}}ois Laviolette, Mario Marchand, and Victor Lempitsky.
\newblock Domain-adversarial training of neural networks.
\newblock {\em JMLR}, 17(1):2096--2030, 2016.

\bibitem{ghifary2016scatter}
Muhammad Ghifary, David Balduzzi, W~Bastiaan Kleijn, and Mengjie Zhang.
\newblock Scatter component analysis: A unified framework for domain adaptation
  and domain generalization.
\newblock {\em IEEE TPAMI}, 39(7):1414--1430, 2016.

\bibitem{ghifary2015domain}
Muhammad Ghifary, W~Bastiaan Kleijn, Mengjie Zhang, and David Balduzzi.
\newblock Domain generalization for object recognition with multi-task
  autoencoders.
\newblock In {\em ICCV}, 2015.

\bibitem{grill2020bootstrap}
Jean-Bastien Grill, Florian Strub, Florent Altch{\'e}, Corentin Tallec, Pierre
  Richemond, Elena Buchatskaya, Carl Doersch, Bernardo Avila~Pires, Zhaohan
  Guo, Mohammad Gheshlaghi~Azar, et~al.
\newblock Bootstrap your own latent-a new approach to self-supervised learning.
\newblock In {\em NeurIPS}, 2020.

\bibitem{gulrajani2020search}
Ishaan Gulrajani and David Lopez-Paz.
\newblock In search of lost domain generalization.
\newblock In {\em ICLR}, 2021.

\bibitem{harary2022unsupervised}
Sivan Harary, Eli Schwartz, Assaf Arbelle, Peter Staar, Shady Abu-Hussein, Elad
  Amrani, Roei Herzig, Amit Alfassy, Raja Giryes, Hilde Kuehne, et~al.
\newblock Unsupervised domain generalization by learning a bridge across
  domains.
\newblock In {\em CVPR}, 2022.

\bibitem{he2020momentum}
Kaiming He, Haoqi Fan, Yuxin Wu, Saining Xie, and Ross Girshick.
\newblock Momentum contrast for unsupervised visual representation learning.
\newblock In {\em CVPR}, 2020.

\bibitem{he2016deep}
Kaiming He, Xiangyu Zhang, Shaoqing Ren, and Jian Sun.
\newblock Deep residual learning for image recognition.
\newblock In {\em CVPR}, 2016.

\bibitem{hu2020domain}
Shoubo Hu, Kun Zhang, Zhitang Chen, and Laiwan Chan.
\newblock Domain generalization via multidomain discriminant analysis.
\newblock In {\em UAI}, 2020.

\bibitem{huang2017arbitrary}
Xun Huang and Serge Belongie.
\newblock Arbitrary style transfer in real-time with adaptive instance
  normalization.
\newblock In {\em ICCV}, 2017.

\bibitem{huang2020self}
Zeyi Huang, Haohan Wang, Eric~P Xing, and Dong Huang.
\newblock Self-challenging improves cross-domain generalization.
\newblock In {\em ECCV}, 2020.

\bibitem{kim2021selfreg}
Daehee Kim, Youngjun Yoo, Seunghyun Park, Jinkyu Kim, and Jaekoo Lee.
\newblock Selfreg: Self-supervised contrastive regularization for domain
  generalization.
\newblock In {\em ICCV}, 2021.

\bibitem{koh2021wilds}
Pang~Wei Koh, Shiori Sagawa, Henrik Marklund, Sang~Michael Xie, Marvin Zhang,
  Akshay Balsubramani, Weihua Hu, Michihiro Yasunaga, Richard~Lanas Phillips,
  Irena Gao, et~al.
\newblock Wilds: A benchmark of in-the-wild distribution shifts.
\newblock In {\em ICML}, 2021.

\bibitem{krueger2021out}
David Krueger, Ethan Caballero, Joern-Henrik Jacobsen, Amy Zhang, Jonathan
  Binas, Dinghuai Zhang, Remi Le~Priol, and Aaron Courville.
\newblock Out-of-distribution generalization via risk extrapolation (rex).
\newblock In {\em ICML}, 2021.

\bibitem{li2017deeper}
Da Li, Yongxin Yang, Yi-Zhe Song, and Timothy~M Hospedales.
\newblock Deeper, broader and artier domain generalization.
\newblock In {\em ICCV}, 2017.

\bibitem{li2018learning}
Da Li, Yongxin Yang, Yi-Zhe Song, and Timothy~M Hospedales.
\newblock Learning to generalize: Meta-learning for domain generalization.
\newblock In {\em AAAI}, 2018.

\bibitem{li2019episodic}
Da Li, Jianshu Zhang, Yongxin Yang, Cong Liu, Yi-Zhe Song, and Timothy~M
  Hospedales.
\newblock Episodic training for domain generalization.
\newblock In {\em ICCV}, 2019.

\bibitem{li2018domain}
Haoliang Li, Sinno~Jialin Pan, Shiqi Wang, and Alex~C Kot.
\newblock Domain generalization with adversarial feature learning.
\newblock In {\em CVPR}, 2018.

\bibitem{li2021simple}
Pan Li, Da Li, Wei Li, Shaogang Gong, Yanwei Fu, and Timothy~M Hospedales.
\newblock A simple feature augmentation for domain generalization.
\newblock In {\em ICCV}, 2021.

\bibitem{li2022uncertainty}
Xiaotong Li, Yongxing Dai, Yixiao Ge, Jun Liu, Ying Shan, and Ling-Yu Duan.
\newblock Uncertainty modeling for out-of-distribution generalization.
\newblock In {\em ICLR}, 2022.

\bibitem{li2021test}
Yizhuo Li, Miao Hao, Zonglin Di, Nitesh~Bharadwaj Gundavarapu, and Xiaolong
  Wang.
\newblock Test-time personalization with a transformer for human pose
  estimation.
\newblock In {\em NeurIPS}, 2021.

\bibitem{li2018deep}
Ya Li, Xinmei Tian, Mingming Gong, Yajing Liu, Tongliang Liu, Kun Zhang, and
  Dacheng Tao.
\newblock Deep domain generalization via conditional invariant adversarial
  networks.
\newblock In {\em ECCV}, 2018.

\bibitem{li2019feature}
Yiying Li, Yongxin Yang, Wei Zhou, and Timothy Hospedales.
\newblock Feature-critic networks for heterogeneous domain generalization.
\newblock In {\em ICML}, 2019.

\bibitem{liu2021ttt++}
Yuejiang Liu, Parth Kothari, Bastien van Delft, Baptiste Bellot-Gurlet, Taylor
  Mordan, and Alexandre Alahi.
\newblock Ttt++: When does self-supervised test-time training fail or thrive?
\newblock In {\em NeurIPS}, 2021.

\bibitem{muandet2013domain}
Krikamol Muandet, David Balduzzi, and Bernhard Sch{\"o}lkopf.
\newblock Domain generalization via invariant feature representation.
\newblock In {\em ICML}, 2013.

\bibitem{nam2021reducing}
Hyeonseob Nam, HyunJae Lee, Jongchan Park, Wonjun Yoon, and Donggeun Yoo.
\newblock Reducing domain gap by reducing style bias.
\newblock In {\em CVPR}, 2021.

\bibitem{pandey2021generalization}
Prashant Pandey, Mrigank Raman, Sumanth Varambally, and Prathosh Ap.
\newblock Generalization on unseen domains via inference-time label-preserving
  target projections.
\newblock In {\em CVPR}, 2021.

\bibitem{peng2019moment}
Xingchao Peng, Qinxun Bai, Xide Xia, Zijun Huang, Kate Saenko, and Bo Wang.
\newblock Moment matching for multi-source domain adaptation.
\newblock In {\em ICCV}, 2019.

\bibitem{pezeshki2021gradient}
Mohammad Pezeshki, Oumar Kaba, Yoshua Bengio, Aaron~C Courville, Doina Precup,
  and Guillaume Lajoie.
\newblock Gradient starvation: A learning proclivity in neural networks.
\newblock In {\em NeurIPS}, 2021.

\bibitem{rame2021ishr}
Alexandre Rame, Corentin Dancette, and Matthieu Cord.
\newblock Fishr: Invariant gradient variances for out-of-distribution
  generalization.
\newblock In {\em ICML}, 2022.

\bibitem{ruan2021optimal}
Yangjun Ruan, Yann Dubois, and Chris~J Maddison.
\newblock Optimal representations for covariate shift.
\newblock In {\em ICLR}, 2022.

\bibitem{russell2008labelme}
Bryan~C Russell, Antonio Torralba, Kevin~P Murphy, and William~T Freeman.
\newblock Labelme: a database and web-based tool for image annotation.
\newblock {\em IJCV}, 77(1):157--173, 2008.

\bibitem{sagawa2019distributionally}
Shiori Sagawa, Pang~Wei Koh, Tatsunori~B Hashimoto, and Percy Liang.
\newblock Distributionally robust neural networks for group shifts: On the
  importance of regularization for worst-case generalization.
\newblock In {\em ICLR}, 2020.

\bibitem{schneider2020improving}
Steffen Schneider, Evgenia Rusak, Luisa Eck, Oliver Bringmann, Wieland Brendel,
  and Matthias Bethge.
\newblock Improving robustness against common corruptions by covariate shift
  adaptation.
\newblock In {\em NeurIPS}, 2020.

\bibitem{selvaraju2017grad}
Ramprasaath~R Selvaraju, Michael Cogswell, Abhishek Das, Ramakrishna Vedantam,
  Devi Parikh, and Dhruv Batra.
\newblock Grad-cam: Visual explanations from deep networks via gradient-based
  localization.
\newblock In {\em ICCV}, 2017.

\bibitem{shi2021gradient}
Yuge Shi, Jeffrey Seely, Philip~HS Torr, N Siddharth, Awni Hannun, Nicolas
  Usunier, and Gabriel Synnaeve.
\newblock Gradient matching for domain generalization.
\newblock In {\em ICLR}, 2021.

\bibitem{sun2016deep}
Baochen Sun and Kate Saenko.
\newblock Deep coral: Correlation alignment for deep domain adaptation.
\newblock In {\em ECCV}, 2016.

\bibitem{sun2020test}
Yu Sun, Xiaolong Wang, Zhuang Liu, John Miller, Alexei Efros, and Moritz Hardt.
\newblock Test-time training with self-supervision for generalization under
  distribution shifts.
\newblock In {\em ICML}, 2020.

\bibitem{vapnik1999nature}
Vladimir Vapnik.
\newblock {\em The nature of statistical learning theory}.
\newblock Springer science \& business media, 1999.

\bibitem{venkateswara2017deep}
Hemanth Venkateswara, Jose Eusebio, Shayok Chakraborty, and Sethuraman
  Panchanathan.
\newblock Deep hashing network for unsupervised domain adaptation.
\newblock In {\em CVPR}, 2017.

\bibitem{wang2020tent}
Dequan Wang, Evan Shelhamer, Shaoteng Liu, Bruno Olshausen, and Trevor Darrell.
\newblock Tent: Fully test-time adaptation by entropy minimization.
\newblock In {\em ICLR}, 2021.

\bibitem{xiao2010sun}
Jianxiong Xiao, James Hays, Krista~A Ehinger, Aude Oliva, and Antonio Torralba.
\newblock Sun database: Large-scale scene recognition from abbey to zoo.
\newblock In {\em CVPR}, 2010.

\bibitem{xiao2022learning}
Zehao Xiao, Xiantong Zhen, Ling Shao, and Cees~GM Snoek.
\newblock Learning to generalize across domains on single test samples.
\newblock In {\em ICLR}, 2022.

\bibitem{xu2021fourier}
Qinwei Xu, Ruipeng Zhang, Ya Zhang, Yanfeng Wang, and Qi Tian.
\newblock A fourier-based framework for domain generalization.
\newblock In {\em CVPR}, 2021.

\bibitem{xu2020robust}
Zhenlin Xu, Deyi Liu, Junlin Yang, Colin Raffel, and Marc Niethammer.
\newblock Robust and generalizable visual representation learning via random
  convolutions.
\newblock In {\em ICLR}, 2021.

\bibitem{yan2020improve}
Shen Yan, Huan Song, Nanxiang Li, Lincan Zou, and Liu Ren.
\newblock Improve unsupervised domain adaptation with mixup training.
\newblock {\em arXiv preprint arXiv:2001.00677}, 2020.

\bibitem{yang2021adversarial}
Fu-En Yang, Yuan-Chia Cheng, Zu-Yun Shiau, and Yu-Chiang~Frank Wang.
\newblock Adversarial teacher-student representation learning for domain
  generalization.
\newblock In {\em NeurIPS}, 2021.

\bibitem{ye2022ood}
Nanyang Ye, Kaican Li, Haoyue Bai, Runpeng Yu, Lanqing Hong, Fengwei Zhou,
  Zhenguo Li, and Jun Zhu.
\newblock Ood-bench: Quantifying and understanding two dimensions of
  out-of-distribution generalization.
\newblock In {\em CVPR}, 2022.

\bibitem{you2021test}
Fuming You, Jingjing Li, and Zhou Zhao.
\newblock Test-time batch statistics calibration for covariate shift.
\newblock {\em arXiv preprint arXiv:2110.04065}, 2021.

\bibitem{zhang2020adaptive}
Marvin Zhang, Henrik Marklund, Nikita Dhawan, Abhishek Gupta, Sergey Levine,
  and Chelsea Finn.
\newblock Adaptive risk minimization: A meta-learning approach for tackling
  group distribution shift.
\newblock {\em arXiv preprint arXiv:2007.02931}, 2020.

\bibitem{zhang2021adaptive}
Marvin Zhang, Henrik Marklund, Nikita Dhawan, Abhishek Gupta, Sergey Levine,
  and Chelsea Finn.
\newblock Adaptive risk minimization: Learning to adapt to domain shift.
\newblock {\em NeurIPS}, 2021.

\bibitem{zhong2022meta}
Tao Zhong, Zhixiang Chi, Li Gu, Yang Wang, Yuanhao Yu, and Jin Tang.
\newblock Meta-dmoe: Adapting to domain shift by meta-distillation from
  mixture-of-experts.
\newblock In {\em NeurIPS}, 2022.

\bibitem{zhou2021domain}
Kaiyang Zhou, Yongxin Yang, Yu Qiao, and Tao Xiang.
\newblock Domain generalization with mixstyle.
\newblock In {\em ICLR}, 2021.

\end{thebibliography}
}

\newpage
\section*{Appendix}

In this supplementary material, we provide,

1. Resource usage for ITTA in Section~\ref{sec resource}.

2. Grad-CAM visualizations of different loss terms in Section~\ref{sec 1}.

3. Parameter analysis of ITTA in Section~\ref{sec 2};

4. Using a different augmentation skill for ITTA in Section~\ref{sec 3}.

5. Using different updating steps or a strategy for ITTA during the test phase in Section~\ref{sec 4}.

6. Using different network structures for the learnable consistency loss and adaptive parameters in Section~\ref{sec 5}.

7. Comparisons with other related methods in Section~\ref{sec 6}. 

8. Detailed experimental results in the DomainBed benchmark in Section~\ref{sec 7}.

\section{Resource Usage Comparisons Between ITTA and the Baseline Model}
\label{sec resource}
Requiring extra resources for our ITTA is a common limitation for existing test-time-based arts. To further evaluate our method, in this section, we compare FLOPS, model size, and inference time in Table~\ref{tab resources}. We compare only with ERM as most existing methods utilize the same network during inferences. We note that compare to the baseline model, ITTA requires extra Flops and processing time, this is because the adaptation process uses extra forward and backward steps during the test phase. While the parameters between the two models are similar because the newly included adaptive blocks are much smaller in size compared to the original model.

\begin{table}[H]
\centering
\caption{Resource comparisons during testing. Here inc. and exc. columns in ITTA indicate to include and exclude the TTA phase.}
\scalebox{0.9}{
\begin{tabular}{lccc}
\toprule 
 Model & Flops (G) &Params (M) &Time (s)\\
 \hline \hline
Baseline &1.82 &11.18 &0.004\\
ITTA (inc. $\vert$ exc.) &~~6.12 $\vert$ 1.83 &~~14.95 $\vert$ 14.94 &~~0.021 $\vert$ 0.005 \\
\bottomrule
\end{tabular}}
\label{tab resources}
\vspace{-0.5 cm}
\end{table}

\section{Grad-CAM Visualizations of Different Self-Supervised Objectives}
\label{sec 1}
In Section~5 of the manuscript, we provide Grad-CAM \cite{selvaraju2017grad} visualizations of our learnable consistency and the main losses to illustrate their alignment. To further show the differences between several TTT tasks \cite{sun2020test,wang2020tent}, we present more visual examples in this section. Results are shown in Figure~\ref{fig 1}. We observe that the entropy minimization \cite{wang2020tent} and rotation estimation \cite{sun2020test} objectives do not activate the same regions as the main loss. As shown in the first row, for the class label of giraffe, both the main loss and our learned loss can correctly locate the two giraffes in the image, while the rotation estimation task can only locate one target, the same observation can be found when the learned weights are disabled in our loss term. Meanwhile, although the two objects can be found for the entropy minimization task, the corresponding hot region does not align with that of the main loss. Similar phenomena can be observed in other samples. These visual examples demonstrate that our learned objective can better align with the main task than the TTT tasks adopted in previous works \cite{sun2020test,wang2020tent}, explaining why using the proposed learnable consistency loss can better improve TTT.

\def\swthree{0.15\linewidth}
\renewcommand{\tabcolsep}{1pt}
\begin{figure*}[t]
\centering
    \begin{tabular}{cccccc}
        \includegraphics[width=\swthree]{image/cams/0/original.pdf}&
        \includegraphics[width=\swthree]{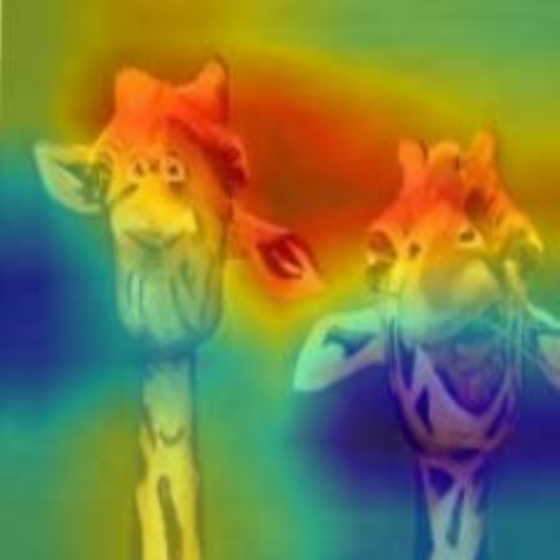}&
        \includegraphics[width=\swthree]{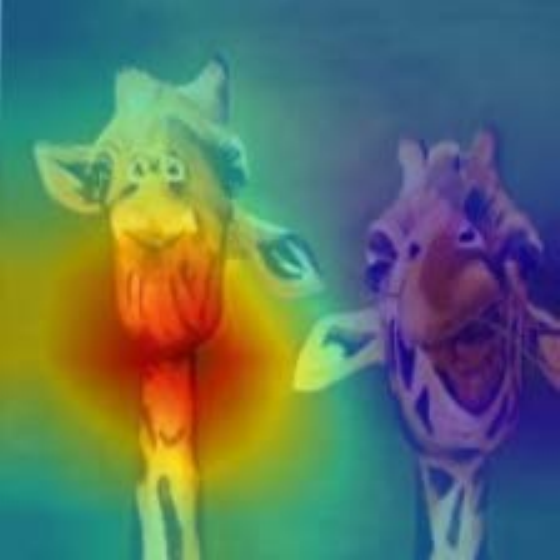}&
        \includegraphics[width=\swthree]{image/cams/0/loss.pdf}&
        \includegraphics[width=\swthree]{image/cams/0/wloss.pdf}&
        \includegraphics[width=\swthree]{image/cams/0/main.pdf}\\
        \includegraphics[width=\swthree]{image/cams/1/original.pdf}&
        \includegraphics[width=\swthree]{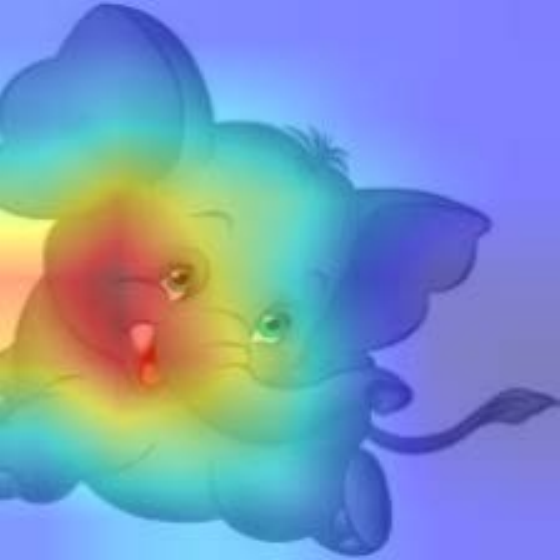}&
        \includegraphics[width=\swthree]{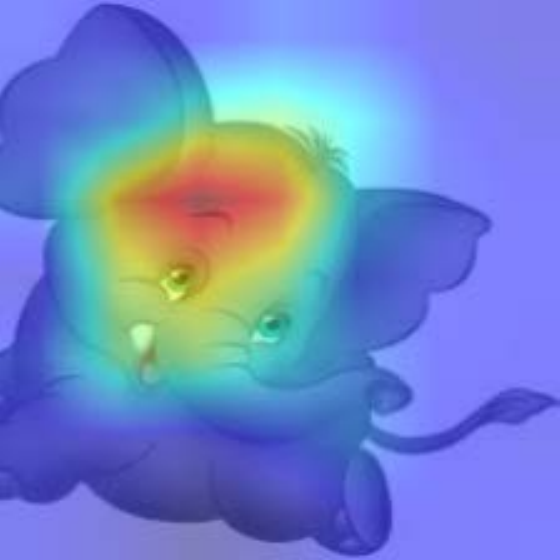}&
        \includegraphics[width=\swthree]{image/cams/1/loss.pdf}&
        \includegraphics[width=\swthree]{image/cams/1/wloss.pdf}&
        \includegraphics[width=\swthree]{image/cams/1/main.pdf}\\
        \includegraphics[width=\swthree]{image/cams/2/original.pdf}&
        \includegraphics[width=\swthree]{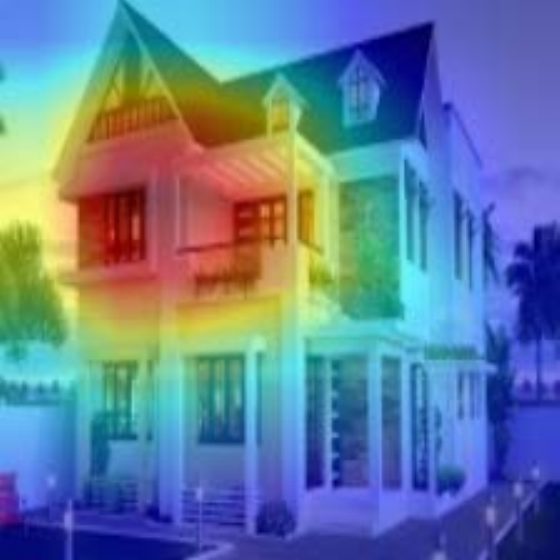}&
        \includegraphics[width=\swthree]{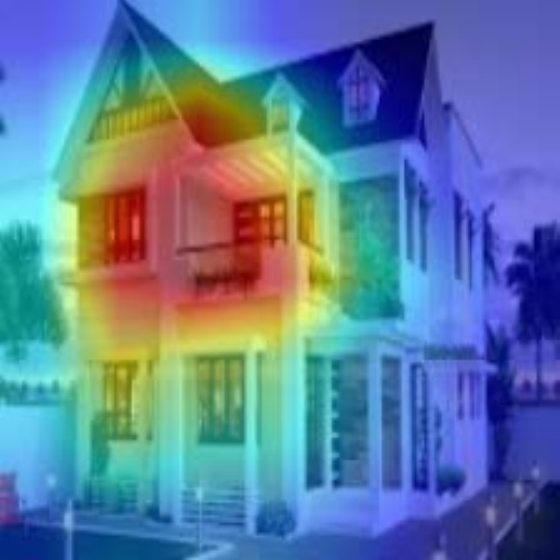}&
        \includegraphics[width=\swthree]{image/cams/2/loss.pdf}&
        \includegraphics[width=\swthree]{image/cams/2/wloss.pdf}&
        \includegraphics[width=\swthree]{image/cams/2/main.pdf}\\
        \includegraphics[width=\swthree]{image/cams/3/original.pdf}&
        \includegraphics[width=\swthree]{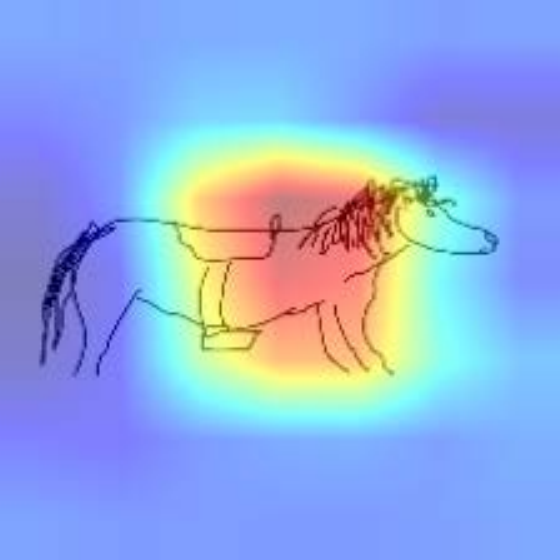}&
        \includegraphics[width=\swthree]{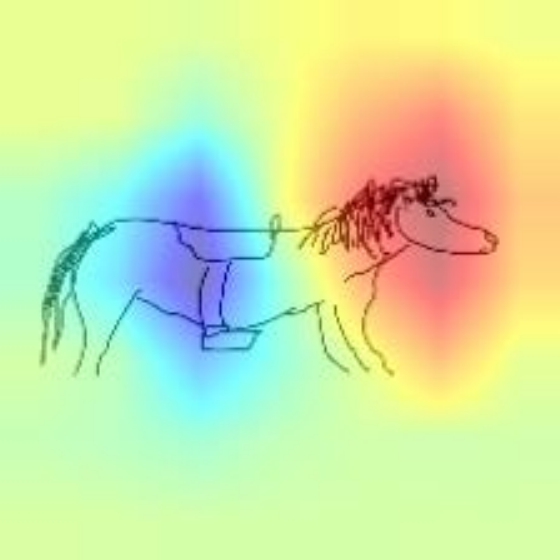}&
        \includegraphics[width=\swthree]{image/cams/3/loss.pdf}&
        \includegraphics[width=\swthree]{image/cams/3/wloss.pdf}&
        \includegraphics[width=\swthree]{image/cams/3/main.pdf}\\
        {\text{(a) Input}} &{\text{(b) Entropy}} &{\text{(c) Rotation}} & {\text{(d) Ours w/o $f_w$}}& {\text{(e) Ours}} & {\text{(f) Main}}\\
    \end{tabular}
    \vspace{-0.2 cm}
	\caption{Grad-CAM \cite{selvaraju2017grad} visualizations from different loss terms. We use images with varying class labels (\ie giraffe, elephant, house, and horse from top to bottom) from the four target domains of PACS \cite{li2017deeper} as inputs (\ie art, cartoon, photo, and sketch domains from top to bottom). ``Entropy" and ``Rotation" here denote the entropy minimization and rotation estimation tasks in \cite{wang2020tent} and \cite{sun2020test}. Ours w/o $f_w$ is the learnable consistency loss in Eq.~(1) in the manuscript (\ie $\Vert f_w(z - z')\Vert$) when $f_w$ is disabled. The proposed learnable consistency loss can align well with the main classification task.}
	\label{fig 1}
\end{figure*}

\section{Parameter Analysis}
\label{sec 2}
In this section, we analyze the hyper-parameter used in ITTA. We use the weight parameter $\alpha$ to balance the contributions from the main loss and weighted consistency loss (\ie $\mathcal{L}_{main} + \alpha\mathcal{L}_{wcont}$ in Eq.~(2) of our manuscript). To analyze the sensitivity of ITTA regarding different values of $\alpha$, we conduct ablation studies in the PACS benchmark \cite{li2017deeper}. Results are listed in Table~\ref{tab app1}. We observe that the proposed ITTA can obtain favorable performances when $\alpha$ is in the range of 0.1 to 10, and it performs the best on average when setting as 1. We thus fix the parameter as 1 in all experiments.

\begin{table*}[h]
\centering
\caption{Sensitivity analysis of ITTA regarding different values of $\alpha$ in the unseen domain from PACS \cite{li2017deeper}. The reported accuracies ($\%$) and standard deviations are computed from 60 trials in each target domain.}
\scalebox{1}{
\begin{tabular}{lC{2cm}C{2cm}C{2cm}C{2cm}C{2cm}}
\toprule 
 \multirow{2}*{Values} & \multicolumn{4}{c}{Target domain} &\multirow{2}*{Avg.}\\
\cline{2-5}
 & Art & Cartoon & Photo & Sketch\\
\hline \hline
$\alpha=0.1$ &83.9 $\pm$ 0.7 &76.2 $\pm$ 1.1 &94.8 $\pm$ 0.2 &78.8 $\pm$ 0.8 &83.4 $\pm$ 0.2\\
$\alpha=1$ (Ours) &84.7 $\pm$ 0.4 &78.0 $\pm$ 0.4 &94.5 $\pm$ 0.4 &78.2 $\pm$ 0.3 &83.8 $\pm$ 0.3 \\
$\alpha=10$ &83.9 $\pm$ 0.5 &77.4 $\pm$ 0.6 &94.2 $\pm$ 0.7 &77.3 $\pm$ 0.8 &83.2 $\pm$ 0.3 \\
$\alpha=100$ &81.5 $\pm$ 1.2 &77.0 $\pm$ 0.6 &92.6 $\pm$ 0.7 &78.9 $\pm$ 2.1 &82.5 $\pm$ 0.9 \\
\bottomrule
\end{tabular}}
\label{tab app1}
\end{table*}

\section{A Different Augmentation Skill for ITTA}
\label{sec 3}
In our manuscript, we use the existing augmentation strategy from \cite{zhou2021domain} to obtain the augmented feature. In this section, we replace this implementation with that from \cite{li2021simple} to further verify if our ITTA can still thrive with another augmentation skill. Different from \cite{zhou2021domain} that mixes the statics of the feature to synthesize new information, \cite{li2021simple} uses an affine transformation to create new features, where the weight for the transformation is sampled from a normal distribution with the mean value of one and standard value of zero, and the bias for the transformation is sampled from a normal distribution with the mean and standard values both zero. Experiments are conducted on the PACS benchmark \cite{li2017deeper} with the leave-one-out strategy.

We compare ITTA with several different variants. (1) Ours w/o $f_w$ \& TTT: this variant is the baseline model which uses the naive consistency loss for training and does not include TTT during the test phase. (2) Ours w/o $f_w$: we disable the $f_w$ in our consistency loss, which uses the naive consistency loss for the test-time updating. (3) Ours w/o TTT: we do not update any parameters during the test phase. This variant is used to verify whether TTT can improve the pretrained model when replacing the augmentation strategy. We also compare these variants with the ERM method to show their effectivenesses.

Results are listed in Table~\ref{tab app2}. We observe that ERM performs favorably against the baseline model, indicating that this augmentation strategy may not be beneficial for the training process. Meanwhile, we observe that when $f_w$ is disabled, the performances seem to decrease in 3 out of 4 target domains, and the average accuracy is also inferior to the baseline (\ie Ours w/o $f_w$ \& TTT). This result is in line with the finding in \cite{liu2021ttt++} that an inappropriate TTT task may deteriorate the performance. In comparison, we note that the performances are both improved when $f_w$ is enabled (\ie Ours w/o TTT and Ours), which once again demonstrates that the proposed learnable consistency loss can improve the trained model. Moreover, we can also observe that when combining $f_w$ and TTT, our model is superior to other variants and the ERM method. These results demonstrate that the proposed two strategies can improve the current TTT framework despite a less effective augmentation strategy.

\begin{table*}[t]
\centering
\caption{Performances of our method with another augmentation strategy from \cite{li2021simple} in the unseen domain from PACS \cite{li2017deeper}. The reported accuracies ($\%$) and standard deviations are computed from 60 trials in each target domain.}
\scalebox{1}{
\begin{tabular}{lC{2cm}C{2cm}C{2cm}C{2cm}C{2cm}}
\toprule 
 \multirow{2}*{Model} & \multicolumn{4}{c}{Target domain} &\multirow{2}*{Avg.}\\
\cline{2-5}
 & Art & Cartoon & Photo & Sketch\\
\hline \hline
ERM &78.0 $\pm$ 1.3 &73.4 $\pm$ 0.8 &94.1 $\pm$ 0.4 &73.6 $\pm$ 2.2 &79.8 $\pm$ 0.4\\
Ours w/o $f_w$ \& TTT &74.9 $\pm$ 0.4 &74.1 $\pm$ 0.8 &90.6 $\pm$ 0.3 &79.7 $\pm$ 0.7 &79.8 $\pm$ 0.4\\
Ours w/o $f_w$ &77.1 $\pm$ 1.0 &73.6 $\pm$ 1.1 &89.9 $\pm$ 0.4 &78.4 $\pm$ 0.8 &79.7 $\pm$ 0.2 \\
Ours w/o TTT &77.5 $\pm$ 0.3 &73.2 $\pm$ 0.6 &92.4 $\pm$ 0.4 &78.0 $\pm$ 1.0 &80.3 $\pm$ 0.3 \\
Ours (w/ $f_w$ \& TTT) &79.2 $\pm$ 0.8  &74.9 $\pm$ 1.1  &92.2 $\pm$ 0.3 &76.9 $\pm$ 0.7 &80.8 $\pm$ 0.4 \\
\bottomrule
\end{tabular}}
\label{tab app2}
\end{table*}

\section{Different Updating Steps or Strategies for ITTA}
\label{sec 4}
In the manuscript, we use one TTT step for ITTA before during the testing step. In this section, we conduct experiments to evaluate the performances of ITTA with different TTT steps. Experiments are conducted on the PACS benchmark \cite{li2017deeper} with the leave-one-out strategy, and each target domain is examined with 60 sets of random seeds and hyper-parameter settings. Results are listed in Table~\ref{tab app3}. We observe that the average accuracies of using more TTT steps are not improved greatly while the computational times are proportional to the TTT steps. To this end, we use one TTT step for ITTA as a compromise between accuracy and efficiency.

We use the online setting from TTT \cite{sun2020test} for all arts, which assumes test samples arrive sequentially and updates the adaptive blocks based on the states optimized from a previous sample. In this section, we also test ITTA in an episodic manner (\ie Epi) \cite{chen2022ost}. Results in Table~\ref{tab app3} suggest that while the episodic updating strategy performs slightly worse than the current scheme, and it still outperforms the baseline.

\begin{table*}[h]
\centering
\caption{Evaluations of ITTA in the unseen domain from PACS \cite{li2017deeper} with different TTT steps and updating strategies during the testing phase. The reported accuracies ($\%$) and standard deviations are computed from 60 trials in each target domain. The time consumption (TC) is computed using one image with the size of 224 $\times$ 224. Epi. denotes updating ITTA in an episodic manner.}
\scalebox{1}{
\begin{tabular}{L{1.8cm}C{2cm}C{2cm}C{2cm}C{2cm}C{2cm}C{2cm}}
\toprule 
 \multirow{2}*{Steps} & \multicolumn{4}{c}{Target domain} &\multirow{2}*{Avg.}&\multirow{2}*{TC}\\
\cline{2-5}
 & Art & Cartoon & Photo & Sketch\\
\hline \hline
1 step (Ours) &84.7 $\pm$ 0.4 &78.0 $\pm$ 0.4 &94.5 $\pm$ 0.4 &78.2 $\pm$ 0.3 &83.8 $\pm$ 0.3 &2.4 ms\\
2 step &84.2 $\pm$ 0.9 &77.5 $\pm$ 0.6 &94.4 $\pm$ 0.4 &79.1 $\pm$ 1.0 &83.8 $\pm$ 0.1 &4.2 ms\\
3 step &84.5 $\pm$ 1.2 &77.6 $\pm$ 0.6 &94.0 $\pm$ 0.6 &79.3 $\pm$ 0.1 &83.9 $\pm$ 0.3 &6.1 ms\\
Epi. &83.6 $\pm$ 0.7 &77.9 $\pm$ 0.5 &95.2 $\pm$ 0.1 &76.6 $\pm$ 0.5 &83.3 $\pm$ 0.4 \\
\bottomrule
\end{tabular}}
\label{tab app3}
\end{table*}

\section{Different Network Structures for the Learnable Consistency Loss and Adaptive Parameters}
\label{sec 5}
In our implementation, we use 10 layers of learnable parameters for $f_w$, and we use 5 layers of learnable parameters for $f_{\Theta}$ after each block. In this section, we evaluate our ITTA with different network structures for these two modules. Specifically, we compare the original implementation with the variants that use 1, 5, and 15 layers for $f_w$ and 1, 10, and 15 layers for $f_{\Theta}$ to evaluate the performances of different structures. Similarly, we conduct experiments on the PACS benchmark \cite{li2017deeper} with the leave-one-out strategy, and each target domain is examined with 60 sets of random seeds and hyper-parameter settings.
Evaluation results are listed in Table~\ref{tab app4}. We observe that their differences in the average accuracy are rather subtle on account of the variances. To this end, we use the original implementation with 10 layers of learnable parameters for $f_w$ and 5 layers of learnable parameters for $f_{\Theta}$, which performs relatively better than other variants.

Since the adaptive blocks $f_{\Theta}$ are attached after each layer of the network, one may wonder how the varying locations of the adaptive blocks affect the performance of ITTA. To answer this question, we further conduct experiments by adding the adaptive blocks after different layers of the original network. Denoting as Loc = $la_n$ given the $n$ layers in the original network, we note that the model performs less effectively when the adaptive block is placed after the 1st layer of the network, and using all four adaptive blocks (\ie ours) is more effective than other alternatives.

\section{Comparisons with Other Related Methods}
\label{sec 6}
Apart from the proposed ITTA, some other works also propose to include learnable parameters in their auxiliary losses. Examples include MetaReg \cite{balaji2018metareg} and Feature-Critic \cite{li2019feature} which both suggest using meta-learning to produce more general models. The main difference between these arts and ITTA is that parameters in the auxiliary loss from \cite{balaji2018metareg,li2019feature} are gradually refined by episode training, and they are updated via a gradient alignment step in ITTA (see Sec.~3.1 in the manuscript), which is much simpler. 
In this section, we compare ITTA with these two arts in the PACS dataset \cite{li2017deeper} using the same settings aforementioned. Because MetaReg \cite{balaji2018metareg} does not release codes, we thus directly cite the data from their paper in the comparison. Different from others, the results in \cite{balaji2018metareg} are averaged by 5 trials according to their paper, which is much less than our experimental settings.
Meanwhile, we also compare with TTT++ \cite{liu2021ttt++} which suggests storing the momentum of the features from the source domain and enforcing the similarity between momentums of features from the source and target domains. We use the same setting in Section 5.1 from the manuscript to evaluate TTT++. 
Results are listed in Table~\ref{tab learnloss}. We observe that our method consistently outperforms that from \cite{balaji2018metareg,li2019feature,liu2021ttt++} for both the cases with and without TTT, indicating that the proposed learnable consistency loss and updating method is not only simpler but also more effective than the losses in \cite{balaji2018metareg,li2019feature}.

\begin{table*}[h]
\centering
\caption{Performances of our method with different network structures for the consistency loss (\ie $f_w$) and adaptive parameters (\ie $f_{\Theta}$) in the unseen domain from PACS \cite{li2017deeper}. Here `Loc=$la_n$' locates the adaptive block after the $n$-th layer of the model (`$la_4$' is the last layer). The reported accuracies ($\%$) and standard deviations are computed from 60 trials in each target domain.}
\vspace{-0.3 cm}
\scalebox{0.9}{
\begin{tabular}{lC{3cm}C{2cm}C{2cm}C{2cm}C{2cm}C{2cm}}
\toprule 
 &\multirow{2}*{Structures} & \multicolumn{4}{c}{Target domain} &\multirow{2}*{Avg.}\\
\cline{3-6}
 && Art & Cartoon & Photo & Sketch\\
\hline \hline
\multirow{4}*{Structures of $f_w$} &1 layer &83.5 $\pm$ 1.2 &76.0 $\pm$ 1.0 &95.3 $\pm$ 0.2 &78.7 $\pm$ 1.5  &83.4 $\pm$ 0.4\\
&5 layers &83.7 $\pm$ 0.6 &76.8 $\pm$ 0.9 &94.6 $\pm$ 0.3  &78.8 $\pm$ 0.3  &83.5 $\pm$ 0.3\\
&10 layers (Ours) &84.7 $\pm$ 0.4 &78.0 $\pm$ 0.4 &94.5 $\pm$ 0.4 &78.2 $\pm$ 0.3 &83.8 $\pm$ 0.3 \\
&15 layers &84.1 $\pm$ 0.4 &75.8 $\pm$ 0.2  &94.3 $\pm$ 0.3 &79.5 $\pm$ 0.4 &83.4 $\pm$ 0.2 \\
\hline
\multirow{4}*{Structures of $f_{\Theta}$} &1 layer &84.0 $\pm$ 0.6 &77.4 $\pm$ 0.5 &94.4 $\pm$ 0.5  &78.3 $\pm$ 0.4  &83.5 $\pm$ 0.3\\
&5 layers (Ours) &84.7 $\pm$ 0.4 &78.0 $\pm$ 0.4 &94.5 $\pm$ 0.4 &78.2 $\pm$ 0.3 &83.8 $\pm$ 0.3\\
&10 layers &84.8 $\pm$ 0.3 &76.0 $\pm$ 0.6 &94.1 $\pm$ 0.5 &78.3 $\pm$ 0.1 &83.3 $\pm$ 0.3 \\
&15 layers &83.9 $\pm$ 0.8 &76.0 $\pm$ 0.5 &93.8 $\pm$ 0.4 &78.7 $\pm$ 1.4 &83.1 $\pm$ 0.6 \\
\hline
\multirow{4}*{Locations of $f_{\Theta}$} &Loc=$la_1$ &83.4 $\pm$ 0.7 &76.8 $\pm$ 0.3 &94.4 $\pm$ 0.3 &77.8 $\pm$ 0.3 &83.1 $\pm$ 0.3 \\
&Loc=$la_2$ &83.4 $\pm$ 0.6 &77.7 $\pm$ 0.6 &94.2 $\pm$ 0.5 &78.0 $\pm$ 0.5 &83.3 $\pm$ 0.3 \\
&Loc=$la_3$ &84.0 $\pm$ 0.4 &77.5 $\pm$ 0.3 &94.4 $\pm$ 0.1 &77.8 $\pm$ 0.1 &83.4 $\pm$ 0.2 \\
&Loc=$la_4$ &84.1 $\pm$ 0.7 &77.8 $\pm$ 0.5 &94.8 $\pm$ 0.2 &76.9 $\pm$ 1.5 &83.4 $\pm$ 0.4 \\
\bottomrule
\end{tabular}}
\label{tab app4}
\end{table*}

\begin{table*}[h]
\centering
\caption{Compare with learnable losses in \cite{balaji2018metareg,li2019feature} in the unseen domain from PACS \cite{li2017deeper}. The reported accuracies ($\%$) and standard deviations are computed from 60 trials in each target domain except for \cite{balaji2018metareg} where the numbers are directly cited from their paper.}
\vspace{-0.3 cm}
\scalebox{1}{
\begin{tabular}{lC{2cm}C{2cm}C{2cm}C{2cm}C{2cm}}
\toprule 
 \multirow{2}*{Model} & \multicolumn{4}{c}{Target domain} &\multirow{2}*{Avg.}\\
\cline{2-5}
 & Art & Cartoon & Photo & Sketch\\
\hline \hline
MetaReg \cite{balaji2018metareg} &83.7 $\pm$ 0.2 &77.2 $\pm$ 0.3 &95.5 $\pm$ 0.2 &70.3 $\pm$ 0.3 &81.7\\
Feture-Critic \cite{li2019feature} &78.4 $\pm$ 1.6  &75.4 $\pm$ 1.2  &92.6 $\pm$ 0.5 &73.3 $\pm$ 1.4 &80.0 $\pm$ 0.3 \\
TTT++ \cite{liu2021ttt++} &84.3 $\pm$ 0.1 &78.4 $\pm$ 0.5 &93.8 $\pm$ 1.3 &73.2 $\pm$ 3.2 &82.4 $\pm$ 1.1 \\
Ours w/o TTT &83.3 $\pm$ 0.5   &76.0 $\pm$ 0.5 &94.4 $\pm$ 0.5  &76.7 $\pm$ 1.4 &82.8 $\pm$ 0.3\\
Ours &84.7 $\pm$ 0.4 &78.0 $\pm$ 0.4 &94.5 $\pm$ 0.4 &78.2 $\pm$ 0.3 &83.8 $\pm$ 0.3 \\
\bottomrule
\end{tabular}}
\label{tab learnloss}
\end{table*}

\section{Detailed Results in the DomainBed Benchmark \cite{gulrajani2020search}}
\label{sec 7}
this section presents the average accuracy in each domain from different datasets. As shown in Table~\ref{tab pacs}, ~\ref{tab vlcs}, ~\ref{tab office}, ~\ref{tab terainc}, and ~\ref{tab domainnet}, these results are detailed illustrations of the results in Table 2 in our manuscript. For all the experiments, we use the ``training-domain validate set" as the model selection method. A total of 22 methods are examined for 60 trials in each unseen domain, and all methods are trained with the leave-one-out strategy using the ResNet18~\cite{he2016deep} backbones.

\begin{table*}[h]
\centering
\caption{Average accuracies on the PACS \cite{li2017deeper} datasets using the default hyper-parameter settings in DomainBed \cite{gulrajani2020search}.}
\vspace{-0.3 cm}
\scalebox{1}{
\begin{tabular}{lC{2cm}C{2cm}C{2cm}C{2cm}C{2cm}}
\toprule 
& art & cartoon & photo & sketch & Average\\
\hline \hline
ERM \cite{vapnik1999nature} &78.0 $\pm$ 1.3 &73.4 $\pm$ 0.8 &94.1 $\pm$ 0.4 &73.6 $\pm$ 2.2 &79.8 $\pm$ 0.4 \\
IRM \cite{arjovsky2019invariant} &76.9 $\pm$ 2.6 &75.1 $\pm$ 0.7 &94.3 $\pm$ 0.4 &77.4 $\pm$ 0.4 &80.9 $\pm$ 0.5\\
GroupGRO \cite{sagawa2019distributionally} &77.7 $\pm$ 2.6 &76.4 $\pm$ 0.3 &94.0 $\pm$ 0.3 &74.8 $\pm$ 1.3 &80.7 $\pm$ 0.4 \\
Mixup \cite{yan2020improve} &79.3 $\pm$ 1.1 &74.2 $\pm$ 0.3 &94.9 $\pm$ 0.3 &68.3 $\pm$ 2.7 &79.2 $\pm$ 0.9 \\
MLDG \cite{li2018learning} &78.4 $\pm$ 0.7 &75.1 $\pm$ 0.5 &94.8 $\pm$ 0.4 &76.7 $\pm$ 0.8 &81.3 $\pm$ 0.2 \\
CORAL \cite{sun2016deep} &81.5 $\pm$ 0.5 &75.4 $\pm$ 0.7 &95.2 $\pm$ 0.5 &74.8 $\pm$ 0.4 &81.7 $\pm$ 0.0 \\
MMD \cite{li2018domain} &81.3 $\pm$ 0.6 &75.5 $\pm$ 1.0 &94.0 $\pm$ 0.5 &74.3 $\pm$ 1.5 &81.3 $\pm$ 0.8 \\
DANN \cite{ganin2016domain} &79.0 $\pm$ 0.6 &72.5 $\pm$ 0.7 &94.4 $\pm$ 0.5 &70.8 $\pm$ 3.0 &79.2 $\pm$ 0.3 \\
CDANN \cite{li2018deep} &80.4 $\pm$ 0.8 &73.7 $\pm$ 0.3 &93.1 $\pm$ 0.6 &74.2 $\pm$ 1.7 &80.3 $\pm$ 0.5 \\
MTL \cite{blanchard2017domain} &78.7 $\pm$ 0.6 &73.4 $\pm$ 1.0 &94.1 $\pm$ 0.6 &74.4 $\pm$ 3.0 &80.1 $\pm$ 0.8 \\
SagNet \cite{nam2021reducing} &82.9 $\pm$ 0.4 &73.2 $\pm$ 1.1 &94.6 $\pm$ 0.5 &76.1 $\pm$ 1.8 &81.7 $\pm$ 0.6 \\
ARM \cite{zhang2020adaptive} &79.4 $\pm$ 0.6 &75.0 $\pm$ 0.7 &94.3 $\pm$ 0.6 &73.8 $\pm$ 0.6 &80.6 $\pm$ 0.5 \\
VREx \cite {krueger2021out} &74.4 $\pm$ 0.7 &75.0 $\pm$ 0.4 &93.3 $\pm$ 0.3 &78.1 $\pm$ 0.9 &80.2 $\pm$ 0.5 \\
RSC \cite{huang2020self} &78.5 $\pm$ 1.1 &73.3 $\pm$ 0.9 &93.6 $\pm$ 0.6 &76.5 $\pm$ 1.4 &80.5 $\pm$ 0.2 \\
SelfReg \cite{kim2021selfreg} &82.5 $\pm$ 0.8 &74.4 $\pm$ 1.5 &95.4 $\pm$ 0.5 &74.9 $\pm$ 1.3 &81.8 $\pm$ 0.3 \\
MixStyle \cite{zhou2021domain} &82.6 $\pm$ 1.2 &76.3 $\pm$ 0.4 &94.2 $\pm$ 0.3 &77.5 $\pm$ 1.3 &82.6 $\pm$ 0.4 \\
Fish \cite{shi2021gradient} &80.9 $\pm$ 1.0 &75.9 $\pm$ 0.4 &95.0 $\pm$ 0.4 & 76.2 $\pm$ 1.0 &82.0 $\pm$ 0.3 \\
SD \cite{pezeshki2021gradient} &83.2 $\pm$ 0.6 &74.6 $\pm$ 0.3 &94.6 $\pm$ 0.1 &75.1 $\pm$ 1.6 &81.9 $\pm$ 0.3 \\
CAD \cite{ruan2021optimal} &83.9 $\pm$ 0.8 &74.2 $\pm$ 0.4 &94.6 $\pm$ 0.4 &75.0 $\pm$ 1.2 &81.9 $\pm$ 0.3 \\
CondCAD \cite{ruan2021optimal} &79.7 $\pm$ 1.0 &74.2 $\pm$ 0.9 &94.6 $\pm$ 0.4 &74.8 $\pm$ 1.4 &80.8 $\pm$ 0.5 \\
Fishr \cite{rame2021ishr} &81.2 $\pm$ 0.4 &75.8 $\pm$ 0.8 &94.3 $\pm$ 0.3 &73.8 $\pm$ 0.6 &81.3 $\pm$ 0.3\\
Ours &84.7 $\pm$ 0.4 &78.0 $\pm$ 0.4 &94.5 $\pm$ 0.4 &78.2 $\pm$ 0.3 &83.8 $\pm$ 0.3 \\
\bottomrule
\end{tabular}}
\label{tab pacs}
\end{table*}

\begin{table*}[h]
\centering
\caption{Average accuracies on the VLCS \cite{fang2013unbiased} datasets using the default hyper-parameter settings in DomainBed \cite{gulrajani2020search}.}
\scalebox{1}{
\begin{tabular}{lC{2cm}C{2cm}C{2cm}C{2cm}C{2cm}}
\toprule 
&Caltech  & LabelMe &Sun & VOC & Average\\
\hline \hline
ERM \cite{vapnik1999nature} &97.7 $\pm$ 0.3 &62.1 $\pm$ 0.9 &70.3 $\pm$ 0.9 &73.2 $\pm$ 0.7 &75.8 $\pm$ 0.2 \\
IRM \cite{arjovsky2019invariant} &96.1 $\pm$ 0.8 &62.5 $\pm$ 0.3 &69.9 $\pm$ 0.7 &72.0 $\pm$ 1.4 &75.1 $\pm$ 0.1\\
GroupGRO \cite{sagawa2019distributionally} &96.7 $\pm$ 0.6 &61.7 $\pm$ 1.5 &70.2 $\pm$ 1.8 &72.9 $\pm$ 0.6 &75.4 $\pm$ 1.0 \\
Mixup \cite{yan2020improve} &95.6 $\pm$ 1.5 &62.7 $\pm$ 0.4 &71.3 $\pm$ 0.3 &75.4 $\pm$ 0.2 &76.2 $\pm$ 0.3 \\
MLDG \cite{li2018learning} &95.8 $\pm$ 0.5 &63.3 $\pm$ 0.8 &68.5 $\pm$ 0.5 &73.1 $\pm$ 0.8 &75.2 $\pm$ 0.3 \\
CORAL \cite{sun2016deep} &96.5 $\pm$ 0.3 &62.8 $\pm$ 0.1 &69.1 $\pm$ 0.6 &73.8 $\pm$ 1.0 &75.5 $\pm$ 0.4 \\
MMD \cite{li2018domain} &96.0 $\pm$ 0.8 &64.3 $\pm$ 0.6 &68.5 $\pm$ 0.6 &70.8 $\pm$ 0.1 &74.9 $\pm$ 0.5 \\
DANN \cite{ganin2016domain} &97.2 $\pm$ 0.1 &63.3 $\pm$ 0.6 &70.2 $\pm$ 0.9 &74.4 $\pm$ 0.2 &76.3 $\pm$ 0.2 \\
CDANN \cite{li2018deep} &95.4 $\pm$ 1.2 &62.6 $\pm$ 0.6 &69.9 $\pm$ 1.3 &76.2 $\pm$ 0.5 &76.0 $\pm$ 0.5 \\
MTL \cite{blanchard2017domain} &94.4 $\pm$ 2.3 &65.0 $\pm$ 0.6 &69.6 $\pm$ 0.6 &71.7 $\pm$ 1.3 &75.2 $\pm$ 0.3 \\
SagNet \cite{nam2021reducing} &94.9 $\pm$ 0.7 &61.9 $\pm$ 0.7 &69.6 $\pm$ 1.3 &75.2 $\pm$ 0.6 &75.4 $\pm$ 0.8 \\
ARM \cite{zhang2020adaptive} &96.9 $\pm$ 0.5 &61.9 $\pm$ 0.4 &71.6 $\pm$ 0.1 &73.3 $\pm$ 0.4 &75.9 $\pm$ 0.3 \\
VREx \cite {krueger2021out} &96.2 $\pm$ 0.0 &62.5 $\pm$ 1.3 &69.3 $\pm$ 0.9 &73.1 $\pm$ 1.2 &75.3 $\pm$ 0.6 \\
RSC \cite{huang2020self} &96.2 $\pm$ 0.0 &63.6 $\pm$ 1.3 &69.8 $\pm$ 1.0 &72.0 $\pm$ 0.4 &75.4 $\pm$ 0.3 \\
SelfReg \cite{kim2021selfreg} &95.8 $\pm$ 0.6 &63.4 $\pm$ 1.1 &71.1 $\pm$ 0.6 &75.3 $\pm$ 0.6 &76.4 $\pm$ 0.7 \\
MixStyle \cite{zhou2021domain} &97.3 $\pm$ 0.3 &61.6 $\pm$ 0.1 &70.4 $\pm$ 0.7 &71.3 $\pm$ 1.9 &75.2 $\pm$ 0.7 \\
Fish \cite{shi2021gradient} &97.4 $\pm$ 0.2 &63.4 $\pm$ 0.1 &71.5 $\pm$ 0.4 &75.2 $\pm$ 0.7 &76.9 $\pm$ 0.2 \\
SD \cite{pezeshki2021gradient} &96.5 $\pm$ 0.4 &62.2 $\pm$ 0.0 &69.7 $\pm$ 0.9 &73.6 $\pm$ 0.4 &75.5 $\pm$ 0.4 \\
CAD \cite{ruan2021optimal} &94.5 $\pm$ 0.9 &63.5 $\pm$ 0.6 &70.4 $\pm$ 1.2 &72.4 $\pm$ 1.3 &75.2 $\pm$ 0.6 \\
CondCAD \cite{ruan2021optimal} &96.5 $\pm$ 0.8 &62.6 $\pm$ 0.4 &69.1 $\pm$ 0.2 &76.0 $\pm$ 0.2 &76.1 $\pm$ 0.3 \\
Fishr \cite{rame2021ishr} &97.2 $\pm$ 0.6 &63.3 $\pm$ 0.7 &70.4 $\pm$ 0.6 &74.0 $\pm$ 0.8 &76.2 $\pm$ 0.3\\
Ours &96.9 $\pm$ 1.2 &63.7 $\pm$ 1.1 &72.0 $\pm$ 0.3 &74.9 $\pm$ 0.8  &76.9 $\pm$ 0.6 \\
\bottomrule
\end{tabular}}
\label{tab vlcs}
\end{table*}

\begin{table*}
\centering
\caption{Average accuracies on the OfficeHome \cite{venkateswara2017deep} datasets using the default hyper-parameter settings in DomainBed \cite{gulrajani2020search}.}
\scalebox{1}{
\begin{tabular}{lC{2cm}C{2cm}C{2cm}C{2cm}C{2cm}}
\toprule 
&art  & clipart &product & real & Average\\
\hline \hline
ERM \cite{vapnik1999nature} &52.2 $\pm$ 0.2 &48.7 $\pm$ 0.5 &69.9 $\pm$ 0.5 &71.7 $\pm$ 0.5 &60.6 $\pm$ 0.2 \\
IRM \cite{arjovsky2019invariant} &49.7 $\pm$ 0.2 &46.8 $\pm$ 0.5 &67.5 $\pm$ 0.4 &68.1 $\pm$ 0.6 &58.0 $\pm$ 0.1\\
GroupGRO \cite{sagawa2019distributionally} &52.6 $\pm$ 1.1 &48.2 $\pm$ 0.9 &69.9 $\pm$ 0.4 &71.5 $\pm$ 0.8 &60.6 $\pm$ 0.3 \\
Mixup \cite{yan2020improve} &54.0 $\pm$ 0.7 &49.3 $\pm$ 0.7 &70.7 $\pm$ 0.7 &72.6 $\pm$ 0.3 &61.7 $\pm$ 0.5 \\
MLDG \cite{li2018learning} &53.1 $\pm$ 0.3 &48.4 $\pm$ 0.3 &70.5 $\pm$ 0.7 &71.7 $\pm$ 0.4 &60.9 $\pm$ 0.2 \\
CORAL \cite{sun2016deep} &55.1 $\pm$ 0.7 &49.7 $\pm$ 0.9 &71.8 $\pm$ 0.2 &73.1 $\pm$ 0.5 &62.4 $\pm$ 0.4 \\
MMD \cite{li2018domain} &50.9 $\pm$ 1.0 &48.7 $\pm$ 0.3 &69.3 $\pm$ 0.7 &70.7 $\pm$ 1.3 &59.9 $\pm$ 0.4 \\
DANN \cite{ganin2016domain} &51.8 $\pm$ 0.5 &47.1 $\pm$ 0.1 &69.1 $\pm$ 0.7 &70.2 $\pm$ 0.7 &59.5 $\pm$ 0.5 \\
CDANN \cite{li2018deep} &51.4 $\pm$ 0.5 &46.9 $\pm$ 0.6 &68.4 $\pm$ 0.5 &70.4 $\pm$ 0.4 &59.3 $\pm$ 0.4 \\
MTL \cite{blanchard2017domain} &51.6 $\pm$ 1.5 &47.7 $\pm$ 0.5 &69.1 $\pm$ 0.3 &71.0 $\pm$ 0.6 &59.9 $\pm$ 0.5 \\
SagNet \cite{nam2021reducing} &55.3 $\pm$ 0.4 &49.6 $\pm$ 0.2 &72.1 $\pm$ 0.4 &73.2 $\pm$ 0.4 &62.5 $\pm$ 0.3 \\
ARM \cite{zhang2020adaptive} &51.3 $\pm$ 0.9 &48.5 $\pm$ 0.4 &68.0 $\pm$ 0.3 &70.6 $\pm$ 0.1 &59.6 $\pm$ 0.3 \\
VREx \cite {krueger2021out} &51.1 $\pm$ 0.3 &47.4 $\pm$ 0.6 &69.0 $\pm$ 0.4 &70.5 $\pm$ 0.4 &59.5 $\pm$ 0.1 \\
RSC \cite{huang2020self} &49.0 $\pm$ 0.1 &46.2 $\pm$ 1.5 &67.8 $\pm$ 0.7 &70.6 $\pm$ 0.3 &58.4 $\pm$ 0.6 \\
SelfReg \cite{kim2021selfreg} &55.1 $\pm$ 0.8 &49.2 $\pm$ 0.6 &72.2 $\pm$ 0.3 &73.0 $\pm$ 0.3 &62.4 $\pm$ 0.1 \\
MixStyle \cite{zhou2021domain} &50.8 $\pm$ 0.6 &51.4 $\pm$ 1.1 &67.6 $\pm$ 1.3 &68.8 $\pm$ 0.5 &59.6 $\pm$ 0.8 \\
Fish \cite{shi2021gradient} &54.6 $\pm$ 1.0 &49.6 $\pm$ 1.0 &71.3 $\pm$ 0.6 &72.4 $\pm$ 0.2 &62.0 $\pm$ 0.6 \\
SD \cite{pezeshki2021gradient} &55.0 $\pm$ 0.4 &51.3 $\pm$ 0.5 &72.5 $\pm$ 0.2 &72.7 $\pm$ 0.3 &62.9 $\pm$ 0.2 \\
CAD \cite{ruan2021optimal} &52.1 $\pm$ 0.6 &48.3 $\pm$ 0.5 &69.7 $\pm$ 0.3 &71.9 $\pm$ 0.4 &60.5 $\pm$ 0.3 \\
CondCAD \cite{ruan2021optimal} &53.3 $\pm$ 0.6 &48.4 $\pm$ 0.2 &69.8 $\pm$ 0.9 &72.6 $\pm$ 0.1 &61.0 $\pm$ 0.4 \\
Fishr \cite{rame2021ishr} &52.6 $\pm$ 0.9 &48.6 $\pm$ 0.3 &69.9 $\pm$ 0.6 &72.4 $\pm$ 0.4 &60.9 $\pm$ 0.3 \\
Ours &54.4 $\pm$ 0.2 &52.3 $\pm$ 0.8 &69.5 $\pm$ 0.3 &71.7 $\pm$ 0.2 &62.0 $\pm$ 0.2 \\
\bottomrule
\end{tabular}}
\label{tab office}
\end{table*}

\newpage
\begin{table*}
\centering
\caption{Average accuracies on the TerraInc \cite{beery2018recognition} datasets using the default hyper-parameter settings in DomainBed \cite{gulrajani2020search}.}
\scalebox{1}{
\begin{tabular}{lC{2cm}C{2cm}C{2cm}C{2cm}C{2cm}}
\toprule 
&L100  & L38 &L43 & L46 & Average\\
\hline \hline
ERM \cite{vapnik1999nature} &42.1 $\pm$ 2.5 &30.1 $\pm$ 1.2 &48.9 $\pm$ 0.6 &34.0 $\pm$ 1.1 &38.8 $\pm$ 1.0 \\
IRM \cite{arjovsky2019invariant} &41.8 $\pm$ 1.8 &29.0 $\pm$ 3.6 &49.6 $\pm$ 2.1 &33.1 $\pm$ 1.5 &38.4 $\pm$ 0.9\\
GroupGRO \cite{sagawa2019distributionally} &45.3 $\pm$ 4.6 &36.1 $\pm$ 4.4 &51.0 $\pm$ 0.8 &33.7 $\pm$ 0.9 &41.5 $\pm$ 2.0 \\
Mixup \cite{yan2020improve} &49.4 $\pm$ 2.0 &35.9 $\pm$ 1.8 &53.0 $\pm$ 0.7          &30.0 $\pm$ 0.9 &42.1 $\pm$ 0.7 \\
MLDG \cite{li2018learning} &39.6 $\pm$ 2.3 &33.2 $\pm$ 2.7 &52.4 $\pm$ 0.5          &35.1 $\pm$ 1.5 &40.1 $\pm$ 0.9 \\
CORAL \cite{sun2016deep} &46.7 $\pm$ 3.2 &36.9 $\pm$ 4.3 &49.5 $\pm$ 1.9          &32.5 $\pm$ 0.7 &41.4 $\pm$ 1.8 \\
MMD \cite{li2018domain} &49.1 $\pm$ 1.2 &36.4 $\pm$ 4.8 &50.4 $\pm$ 2.1          &32.3 $\pm$ 1.5 &42.0 $\pm$ 1.0 \\
DANN \cite{ganin2016domain} &44.3 $\pm$ 3.6 &28.0 $\pm$ 1.5 &47.9 $\pm$ 1.0          &31.3 $\pm$ 0.6 &37.9 $\pm$ 0.9 \\
CDANN \cite{li2018deep} &36.9 $\pm$ 6.4 &32.7 $\pm$ 6.2 &51.1 $\pm$ 1.3          &33.5 $\pm$ 0.5 &38.6 $\pm$ 2.3 \\
MTL \cite{blanchard2017domain} &45.2 $\pm$ 2.6 &31.0 $\pm$ 1.6 &50.6 $\pm$ 1.1          &34.9 $\pm$ 0.4 &40.4 $\pm$ 1.0 \\
SagNet \cite{nam2021reducing} &36.3 $\pm$ 4.7 &40.3 $\pm$ 2.0 &52.5 $\pm$ 0.6          &33.3 $\pm$ 1.3 &40.6 $\pm$ 1.5 \\
ARM \cite{zhang2020adaptive} &41.5 $\pm$ 4.5 &27.7 $\pm$ 2.4 &50.9 $\pm$ 1.0          &29.6 $\pm$ 1.5 &37.4 $\pm$ 1.9 \\
VREx \cite {krueger2021out} &48.0 $\pm$ 1.7 &41.1 $\pm$ 1.5 &51.8 $\pm$ 1.5          &32.0 $\pm$ 1.2 &43.2 $\pm$ 0.3 \\
RSC \cite{huang2020self} &42.8 $\pm$ 2.4 &32.2 $\pm$ 3.8 &49.6 $\pm$ 0.9          &32.9 $\pm$ 1.2 &39.4 $\pm$ 1.3 \\
SelfReg \cite{kim2021selfreg} &46.1 $\pm$ 1.5 &34.5 $\pm$ 1.6 &49.8 $\pm$ 0.3          &34.7 $\pm$ 1.5 &41.3 $\pm$ 0.3 \\
MixStyle \cite{zhou2021domain} &50.6 $\pm$ 1.9 &28.0 $\pm$ 4.5 &52.1 $\pm$ 0.7          &33.0 $\pm$ 0.2 &40.9 $\pm$ 1.1 \\
Fish \cite{shi2021gradient} &46.3 $\pm$ 3.0 &29.0 $\pm$ 1.1 &52.7 $\pm$ 1.2          &32.8 $\pm$ 1.0 &40.2 $\pm$ 0.6 \\
SD \cite{pezeshki2021gradient} &45.5 $\pm$ 1.9 &33.2 $\pm$ 3.1 &52.9 $\pm$ 0.7          &36.4 $\pm$ 0.8 &42.0 $\pm$ 1.0 \\
CAD \cite{ruan2021optimal} &43.1 $\pm$ 2.6 &31.1 $\pm$ 1.9 &53.1 $\pm$ 1.6          &34.7 $\pm$ 1.3 &40.5 $\pm$ 0.4 \\
CondCAD \cite{ruan2021optimal} &44.4 $\pm$ 2.9 &32.9 $\pm$ 2.5 &50.5 $\pm$ 1.3          &30.8 $\pm$ 0.5 &39.7 $\pm$ 0.4 \\
Fishr \cite{rame2021ishr} &49.9 $\pm$ 3.3 &36.6 $\pm$ 0.9 &49.8 $\pm$ 0.2 &34.2 $\pm$ 1.3 &42.6 $\pm$ 1.0 \\
Ours &51.7 $\pm$ 2.4 &37.6 $\pm$ 0.6 &49.9 $\pm$ 0.6 &33.6 $\pm$ 0.6 &43.2 $\pm$ 0.5 \\
\bottomrule
\end{tabular}}
\label{tab terainc}
\end{table*}

\begin{table*}
\centering
\caption{Average accuracies on the DomainNet \cite{peng2019moment} datasets using the default hyper-parameter settings in DomainBed  \cite{gulrajani2020search}.}
\scalebox{1}{
\begin{tabular}{lC{2cm}C{2cm}C{2cm}C{2cm}C{2cm}C{2cm}C{2 cm}}
\toprule 
&clip  & info &paint & quick &real &sketch & Average\\
\hline \hline
ERM \cite{vapnik1999nature} &50.4 $\pm$ 0.2 &14.0 $\pm$ 0.2 &40.3 $\pm$ 0.5          &11.7 $\pm$ 0.2 &52.0 $\pm$ 0.2 &43.2 $\pm$ 0.3 &35.3 $\pm$ 0.1 \\
IRM \cite{arjovsky2019invariant} &43.2 $\pm$ 0.9 &12.6 $\pm$ 0.3 &35.0 $\pm$ 1.4          &9.9 $\pm$ 0.4 &43.4 $\pm$ 3.0 &38.4 $\pm$ 0.4 &30.4 $\pm$ 1.0\\
GroupGRO \cite{sagawa2019distributionally} &38.2 $\pm$ 0.5 &13.0 $\pm$ 0.3          &28.7 $\pm$ 0.3 &8.2 $\pm$ 0.1 &43.4 $\pm$ 0.5 &33.7 $\pm$ 0.0 &27.5 $\pm$ 0.1 \\
Mixup \cite{yan2020improve} &48.9 $\pm$ 0.3 &13.6 $\pm$ 0.3 &39.5 $\pm$ 0.5          &10.9 $\pm$ 0.4 &49.9 $\pm$ 0.2 &41.2 $\pm$ 0.2 &34.0 $\pm$ 0.0 \\
MLDG \cite{li2018learning} &51.1 $\pm$ 0.3 &14.1 $\pm$ 0.3 &40.7 $\pm$ 0.3          &11.7 $\pm$ 0.1 &52.3 $\pm$ 0.3 &42.7 $\pm$ 0.2 &35.4 $\pm$ 0.0 \\
CORAL \cite{sun2016deep} &51.2 $\pm$ 0.2 &15.4 $\pm$ 0.2 &42.0 $\pm$ 0.2 &12.7 $\pm$ 0.1 &52.0 $\pm$ 0.3 &43.4 $\pm$ 0.0 &36.1 $\pm$ 0.2 \\
MMD \cite{li2018domain} &16.6 $\pm$ 13.3 &0.3 $\pm$ 0.0 &12.8 $\pm$ 10.4 &0.3 $\pm$ 0.0 &17.1 $\pm$ 13.7 &0.4 $\pm$ 0.0 &7.9 $\pm$ 6.2 \\
DANN \cite{ganin2016domain}  &45.0 $\pm$ 0.2 &12.8 $\pm$ 0.2 &36.0 $\pm$ 0.2          &10.4 $\pm$ 0.3 &46.7 $\pm$ 0.3 &38.0 $\pm$ 0.3 &31.5 $\pm$ 0.1\\
CDANN \cite{li2018deep} &45.3 $\pm$ 0.2 &12.6 $\pm$ 0.2 &36.6 $\pm$ 0.2          &10.3 $\pm$ 0.4 &47.5 $\pm$ 0.1 &38.9 $\pm$ 0.4 &31.8 $\pm$ 0.2 \\
MTL \cite{blanchard2017domain} &50.6 $\pm$ 0.2 &14.0 $\pm$ 0.4 &39.6 $\pm$ 0.3          &12.0 $\pm$ 0.3 &52.1 $\pm$ 0.1 &41.5 $\pm$ 0.0 &35.0 $\pm$ 0.0 \\
SagNet \cite{nam2021reducing} &51.0 $\pm$ 0.1 &14.6 $\pm$ 0.1 &40.2 $\pm$ 0.2          &12.1 $\pm$ 0.2 &51.5 $\pm$ 0.3 &42.4 $\pm$ 0.1 &35.3 $\pm$ 0.1 \\
ARM \cite{zhang2020adaptive} &43.0 $\pm$ 0.2 &11.7 $\pm$ 0.2 &34.6 $\pm$ 0.1          &9.8 $\pm$ 0.4 &43.2 $\pm$ 0.3 &37.0 $\pm$ 0.3 &29.9 $\pm$ 0.1 \\
VREx \cite {krueger2021out} &39.2 $\pm$ 1.6 &11.9 $\pm$ 0.4 &31.2 $\pm$ 1.3          &10.2 $\pm$ 0.4 &41.5 $\pm$ 1.8 &34.8 $\pm$ 0.8 &28.1 $\pm$ 1.0 \\
RSC \cite{huang2020self} &39.5 $\pm$ 3.7 &11.4 $\pm$ 0.8 &30.5 $\pm$ 3.1          &10.2 $\pm$ 0.8 &41.0 $\pm$ 1.4 &34.7 $\pm$ 2.6 &27.9 $\pm$ 2.0 \\
SelfReg \cite{kim2021selfreg} &47.9 $\pm$ 0.3 &15.1 $\pm$ 0.3 &41.2 $\pm$ 0.2          &11.7 $\pm$ 0.3 &48.8 $\pm$ 0.0 &43.8 $\pm$ 0.3 &34.7 $\pm$ 0.2 \\
MixStyle \cite{zhou2021domain} &49.1 $\pm$ 0.4 &13.4 $\pm$ 0.0 &39.3 $\pm$ 0.0 &11.4 $\pm$ 0.4 &47.7 $\pm$ 0.3 &42.7 $\pm$ 0.1 &33.9 $\pm$ 0.1\\
Fish \cite{shi2021gradient} &51.5 $\pm$ 0.3 &14.5 $\pm$ 0.2 &40.4 $\pm$ 0.3          &11.7 $\pm$ 0.5 &52.6 $\pm$ 0.2 &42.1 $\pm$ 0.1 &35.5 $\pm$ 0.0 \\
SD \cite{pezeshki2021gradient} &51.3 $\pm$ 0.3 &15.5 $\pm$ 0.1 &41.5 $\pm$ 0.3          &12.6 $\pm$ 0.2 &52.9 $\pm$ 0.2 &44.0 $\pm$ 0.4 &36.3 $\pm$ 0.2 \\
CAD \cite{ruan2021optimal} &45.4 $\pm$ 1.0 &12.1 $\pm$ 0.5 &34.9 $\pm$ 1.1          &10.2 $\pm$ 0.6 &45.1 $\pm$ 1.6 &38.5 $\pm$ 0.6 &31.0 $\pm$ 0.8 \\
CondCAD \cite{ruan2021optimal} &46.1 $\pm$ 1.0 &13.3 $\pm$ 0.4 &36.1 $\pm$ 1.4          &10.7 $\pm$ 0.2 &46.8 $\pm$ 1.3 &38.7 $\pm$ 0.7 &31.9 $\pm$ 0.7 \\
Fishr \cite{rame2021ishr} &47.8 $\pm$ 0.7 &14.6 $\pm$ 0.2 &40.0 $\pm$ 0.3 &11.9 $\pm$ 0.2 &49.2 $\pm$ 0.7 &41.7 $\pm$ 0.1 &34.2 $\pm$ 0.3 \\
Ours &50.7 $\pm$ 0.7 &13.9 $\pm$ 0.4 &39.4 $\pm$ 0.5 &11.9 $\pm$ 0.2 &50.2 $\pm$ 0.3 &43.5 $\pm$ 0.1 &34.9 $\pm$ 0.1 \\
\bottomrule
\end{tabular}}
\label{tab domainnet}
\end{table*}

\end{document}